%% file: main.tex
\documentclass[runningheads]{llncs}

\usepackage{eccv}
\usepackage{eccvabbrv}
\usepackage[T1]{fontenc}
\usepackage[utf8]{inputenc}
\usepackage{pmboxdraw}
\usepackage{graphicx}
\usepackage{booktabs}
\usepackage[accsupp]{axessibility}  %
\usepackage[colorlinks,citecolor=eccvblue]{hyperref}
\usepackage{orcidlink}
\usepackage{graphicx}
\usepackage{tikz}
\usepackage{comment}
\usepackage{amsmath,amssymb} %
\usepackage{color, colortbl}
\usepackage[numbers,sort&compress]{natbib}

\usepackage[export]{adjustbox} %
\usepackage{booktabs} %

\usepackage{amssymb}
\usepackage{mathtools}
\usepackage{stmaryrd}
\usepackage{tabularx}
\usepackage{makecell}

\usepackage{url}            %
\usepackage{booktabs}       %
\usepackage{amsfonts}       %
\usepackage{nicefrac}       %
\usepackage{microtype}      %
\usepackage{multirow}       %

\usepackage{enumitem}%
\usepackage{wrapfig}
\usepackage{comment}
\usepackage{amsmath,amssymb} %
\usepackage{algorithm}
\usepackage[noend]{algpseudocode}
\usepackage{mathtools}%
\usepackage{arydshln}
\usepackage{symbols}
\usepackage{pifont}

\input{math_commands}

\newcommand*{\ShowNotes}{} %
\input{macros.tex}

\usepackage[multiple]{footmisc}

\newcommand{\myparagraph}[1]{\vspace*{0pt}{\bf #1}}

\definecolor{mybrown}{rgb}{0.87058824, 0.56078431, 0.01960784}
\definecolor{myblue}{rgb}{0.3372549 , 0.70588235, 0.91372549}
\definecolor{mypurple}{rgb}{0.8, 0.47058824, 0.7372549 }
\definecolor{myorange}{rgb}{0.835, 0.368, 0}
\definecolor{mygreen}{rgb}{0.00784314, 0.61960784, 0.45098039}
\definecolor{mygt}{rgb}{0.0078125 , 0.57421875, 0.40625}
\definecolor{mysp}{rgb}{0.84765625, 0.515625  , 0.0234375}
\definecolor{mycitecolor}{rgb}{0,0.08,0.45}
\definecolor{mygr}{rgb}{0.9607,0.9607,0.9607}
\definecolor{myoo}{rgb}{0.992,0.9176,0.9019}

\usepackage{listings}

\begin{document}

\title{
Deep Nets with Subsampling Layers Unwittingly Discard Useful Activations at Test-Time
} 

\titlerunning{Subsampling Layers Unwittingly
Discard Useful Activations}

\author{Chiao-An Yang\inst{1} \and
Ziwei Liu\inst{2} \and
Raymond A. Yeh\inst{1}}

\authorrunning{C.-A. Yang et al.}

\institute{
Department of Computer Science, Purdue University
\\ 
\and
S-Lab, Nanyang Technological University
}

\maketitle

\input{sec00_abs}
\input{sec01_intro}

\input{sec02_rel}

\input{sec03_app}

\input{sec04_exp}

\input{sec05_conc}

\section*{Acknowledgements}
We thank Renan A. Rojas-Gomez and Teck-Yian Lim for their helpful discussions.

\bibliographystyle{splncs04nat}
\bibliography{refs}

\newpage
\appendix
\onecolumn
\input{sec06_appendix}

\end{document}

%% file: math_commands.tex
\usepackage{amsmath,amsfonts,bm}

\def\floor#1{\lfloor #1 \rfloor}
\def\1{\bm{1}}

\def\sp{space}

\def\vzero{{\bm{0}}}

\def\vf{{\bm{f}}}

\def\vk{{\bm{k}}}

\def\vq{{\bm{q}}}

\def\vs{{\bm{s}}}

\def\vv{{\bm{v}}}

\def\vx{{\bm{x}}}
\def\vy{{\bm{y}}}

\def\mF{{\bm{F}}}

\def\mI{{\bm{I}}}

\def\mW{{\bm{W}}}

\DeclareMathAlphabet{\mathsfit}{\encodingdefault}{\sfdefault}{m}{sl}
\SetMathAlphabet{\mathsfit}{bold}{\encodingdefault}{\sfdefault}{bx}{n}

\def\gD{{\mathcal{D}}}

\def\gF{{\mathcal{F}}}

\def\gL{{\mathcal{L}}}

\def\gS{{\mathcal{S}}}

\def\sN{{\mathbb{N}}}

\def\sR{{\mathbb{R}}}

\def\sZ{{\mathbb{Z}}}

%% file: macros.tex
\definecolor{darkred}{rgb}{0.7,0.1,0.1}
\definecolor{darkgreen}{rgb}{0.1,0.7,0.1}
\definecolor{cyan}{rgb}{0.7,0.0,0.7}
\definecolor{dblue}{rgb}{0.2,0.2,0.8}
\definecolor{maroon}{rgb}{0.76,.13,.28}
\definecolor{burntorange}{rgb}{0.81,.33,0}
\definecolor{tealblue}{rgb}{0.212,0.459, 0.533}
\definecolor{myyellow}{rgb}{0.8627451 , 0.75294118, 0.20784314]}
	
\definecolor{LightCyan}{rgb}{0.88,1,1}
\newcommand{\ours}{\cellcolor{LightCyan}}

\definecolor{mypink}{rgb}{0.93359375, 0.62109375, 0.83984375}

\definecolor{pp}{rgb}{0.43921569, 0.18823529, 0.62745098}
\definecolor{rr}{rgb}{0.5254902 , 0.00784314, 0.12941176}
\definecolor{bb}{rgb}{0.09019608, 0.23529412, 0.37647059}
\definecolor{yy}{rgb}{0.49803922, 0.3372549 , 0.0}
\definecolor{gg}{rgb}{0.02352941, 0.3372549 , 0.17647059}

\definecolor{code_bg}{rgb}{0.95, 0.95, 0.95}

\ifdefined\ShowNotes
  \newcommand{\colornote}[3]{{\color{#1}\bf{#2: #3}\normalfont}}
\else
  \newcommand{\colornote}[3]{}
\fi

\definecolor{lightred}{rgb}{0.9,0.4,0.4}

%% file: sec00_abs.tex
\begin{abstract}
Subsampling layers play a crucial role in deep nets by discarding a portion of an activation map to reduce its spatial dimensions. This encourages the deep net to learn higher-level representations. Contrary to this motivation, we hypothesize that the discarded activations are useful and can be incorporated on the fly to improve models' prediction. To validate our hypothesis, we propose a search and aggregate method to find useful activation maps to be used at test time. We applied our approach to the task of image classification and semantic segmentation. Extensive experiments over nine different architectures on multiple datasets show that our method consistently improves model test-time performance, complementing existing test-time augmentation techniques.
Our code is available at \url{https://github.com/ca-joe-yang/discard-in-subsampling}.
\end{abstract}

%% file: sec01_intro.tex
\section{Introduction}
In computer vision, deep nets are commonly trained with the assumption that data samples are drawn independently and identically from an unknown distribution~\cite{murphy2022probabilistic}. Following this assumption, it is intuitive that the same model,~\ie, same forward pass, should be applied to all samples during both the training and testing time. However, when the assumption is not met then \textit{changing the test time procedure} may lead to better performance. 
For example, test-time augmentation (TTA) leverages additional prior information, \ie, knowing the suitable augmentations, over the data distribution to improve model performance.
For vision models, TTA methods apply random augmentations,~\eg, random crops, flips, and rotation, to the test image and perform majority voting to make a final prediction~\cite{szegedy2015going, Simonyan15, he2017mask,pavllo20193d,yang2018deep}. With the success of TTA, a natural question arises: are there other choices for modifying the test-time procedure?

In this work, we present an orthogonal approach to improving the model performance at test-time. Instead of imposing additional knowledge through data augmentation, we re-examine the ones that are \textit{built into the deep net architecture}. We focus on the knowledge built into subsampling and pooling layers. %
Our observation is that models with subsampling layers do not utilize activations to their fullest, as some activations are discarded. %
A question arises, can the discharged activations benefit the model?
The main challenges are: (a) identifying which of the discarded activations are useful; and (b) how to incorporate these activations into a test-time procedure.

We formulate an activations search space for a given pre-trained deep net. Each state in this space corresponds to an activations map that can be extracted by choosing different selection indices ($s$) in the subsampling layers. Given a computation budget, we conduct a greedy search for the set of most promising activation maps based on a confidence criterion. We then aggregate these activation maps using a weighted average to make a final prediction. %

We conduct extensive experiments on ImageNet~\cite{krizhevsky2012imagenet} over nine different pre-trained networks, including ConvNets and Vision Transformers, to validate the efficacy of the proposed approach. We also evaluate our method on the task of semantic segmentation using Cityscapes~\cite{cordts2016Cityscapes} and ADE20K~\cite{zhou2019semantic} with different segmentation networks, including FCN~\cite{long2015fully}, DeepLab~\cite{chen2017rethinking, chen2018encoder}, and SegFormer~\cite{xie2021segformer}. Overall, we find that our test-time procedure improves both classification and segmentation performance. Additionally, our approach achieves \textit{additional gains} when it is used in conjunction with existing TTA methods, \ie, our approach complements TTA.

\myparagraph{\noindent Our contributions are as follows:}
\begin{itemize}[topsep=2pt]
\item We identified that deep nets with subsampling layers discard activations that could be useful for prediction.
\item We propose a framework to search over the discarded activations with a learned criterion and aggregate useful ones, via an attention aggregation module, to improve model performance at test-time. %
\item Extensive experiments on various deep nets demonstrate the effectiveness of the proposed test-time procedure on both image classification and semantic segmentation tasks.
\end{itemize}

%% file: sec02_rel.tex
\section{Related work}
We briefly discuss related work in test-time augmentation, adaptation, ensemble models, and pooling layers.

\myparagraph{Test-time augmentation.} 
As our approach can be viewed as a form of Test-Time Augmentation (TTA) but over the set of selection indices, hence we briefly review TTA.
At a high level, TTA aims to increase a model's performance at test-time by {\bf at the cost of computation increases.} This is beneficial for tasks where the risk of making a mistake significantly outweighs the computation~\eg, medical-related tasks. We note that the computation increase may be very significant. Let's consider image classification on ImageNet, TTA applied to their model includes 144 crops per-image~\cite{szegedy2015going}, \ie, {\bf an increase 144$\times$ in compute.} More recent TTA method~\cite{shanmugam2021better}, further increases the computation to $150\times$.

The most common form of TTA is to augment the input image and combine the output from each augmentation to make a final prediction for improving task performance. Common augmentations include random cropping~\cite{szegedy2015going, Simonyan15, bahat2020classification}, flipping~\cite{he2017mask,pavllo20193d}, or combinations of the augmentations~\cite{moshkov2020test}. See~\figref{fig:overview} for a comparison of standard, test-time augmentation, and our test-time procedure. 

Recent works~\cite{kim2020learning, lyzhov2020greedy, shanmugam2021better, chun2022cyclic, li2023joint, moshkov2020test, gaillochet2022taal} try to learn the distribution of data augmentation for test-time impacts. Given a set of TTA augmentation, a common approach is to average the resulting logits to make a final prediction. GPS~\cite{lyzhov2020greedy} learns to pick the top-$k$ augmentation transforms based on an iterative greedy search.
~\citet{shanmugam2021better} learn a weighted function, deciding which augmentation transform is more important than the other. 
~\citet{chun2022cyclic} also studied how to effectively combine/aggregate augmentations' output into a single prediction via the Entropy Weight Method (EWM)~\cite{liu2010using, amiri2014groundwater},~\ie, the final prediction is a weighted combination based on entropy. DiffTPT~\cite{feng2023diverse} leverages diffusion models to generate high-quality augmented data. TeSLA~\cite{tomar2023tesla} introduced flipped cross-entropy loss to adapt the pre-trained model during test time.~\citet{kim2020learning} learn to select designated data transformation for each image by introducing an auxiliary module on estimating losses.  
Unfortunately, we are not able to reproduce their results~\cite{kim2020learning} due to its heavy computation requirement.

\input{figs/overview}

It is to be noted that our method differs from the aforementioned works in several ways. First, all TTA methods modify the inputs on the image space using various data transformations. On the other hand, we use the discarded activations in the feature space of the subsampling layers. Second, unlike these TTA methods that learn general domain knowledge, our searching and aggregation procedure adapts to \textit{each image} and generates an instance-based augmented feature. Finally, our learned aggregation can be generalized to different test-time budgets \textit{without} the need for retraining. In contrast, existing works~\cite{lyzhov2020greedy, shanmugam2021better} need to retrain their aggregation layer if they change the test-time budget.

\myparagraph{Ensemble methods.}
Classical ensemble methods such as stacking~\cite{wolpert1992stacked} and bagging~\cite{breiman1996bagging} aggregate predictions from multiple models to create an improved prediction. Recent works~\cite{wen2020batchensemble,nam2021diversity,zhang2020diversified,huangsnapshot} focus on how to efficiently learn a set of diverse models. We note that ensemble models require training \textit{multiple models} and running each of them at test time. On the contrary, our approach only requires \textit{a single} pre-trained model (with subsampling layers) to make a prediction. In other words, our approach is orthogonal to the ensemble methods. 
If the ensembled deep nets contain subsampling layers, then we can apply our approach on top of it.

\myparagraph{Subsampling and pooling layers.}
Subsampling layers can be traced back to the origin of convolutional neural networks~\cite{Fukushima_1980,lecun1999object} where striding is used to reduce the spatial dimension and increase the networks' receptive field. Another choice to reduce the spatial dimension is pooling layers, \eg, Average Pooling~\cite{LeCun_Boser_Denker_Henderson_Howard_Hubbard_Jackel_1989}, Max-Pooling~\cite{yamaguchi1990neural,ranzato2007unsupervised}, or other generalizations~\cite{Sermanet_Chintala_LeCun_2012,zeiler2013stochastic,shi2016rank,rojas2022learnable}. 
To preserve the lost information from downsampling layers, anti-aliased CNN~\cite{zhang2019making} inserts a {\tt max-blur-pool} layer to the pooling layer.
In this work, we show that deep nets with subsampling layers can improve their test-time performance by using discarded activations. Additional relevant details and notation of subsampling layers are reviewed in~\secref{sec:prelim}.

\section{Preliminaries}\label{sec:prelim}
We provide a brief review of the subsampling layer and how it is used in deep nets. For readability, we describe these ideas on 1D ``images'', in practice, they are generalized to 2D activation maps with multiple channels.\\
\vspace{-0.35cm}
\begin{wrapfigure}[7]{r}{0.45\linewidth}%
\vspace{-0.6cm}
\input{figs/subsampling}%
\vspace{-0.1cm}
\end{wrapfigure}%
\indent\myparagraph{Subsampling layers.}
In its simplest form, a subsampling operation ${\tt Sub}_R: \sR^N \mapsto \sR^{\floor{N/R}}$ with a subsampling rate $R \in \sN$ reduces an input image $\mI$'s spatial size from $N$ to $\floor{N/R}$ following:
\vspace{-0.1cm}
\bea\label{eq:sub}
{\tt Sub}_R(\mI; s)[n] = \mI[Rn+s]~~\forall n\in\sZ,
\eea
where $s \in \{0, \dots, R-1\}$ denotes a selection index. In most deep learning frameworks~\cite{abadi2016tensorflow,pytorch,jax2018github}, a subsampling layer defaults to $s=0$, \ie, when subsampling by a factor of two ($R=2$) the activations on the even indices (s=0) are kept and the odd indices ($s=1$) activations are discarded; as illustrated in~\figref{fig:subsampling}. In this work, we show that using the discarded activation, \ie, $s\neq 0$, at test-time can lead to performance gains.

\myparagraph{Deep nets with subsampling layers.} Many deep nets in computer vision contain subsampling layers to reduce spatial dimensions. We take a very generic view toward subsampling layers. Consider a convolution layer with stride two, it can be viewed as a convolution layer with stride \textit{one} followed by a subsampling by a factor of two. Similarly, any pooling operation can be decomposed into a ``sliding'' operation followed by subsampling. For example, max pooling is equivalent to a sliding max filter followed by a subsampling layer. With this perspective, many notable deep net architectures contain subsampling layers, including early works such as LeNet~\cite{lecun1999object}, the popular ResNet~\cite{he2016deep}, MobileNet~\cite{howard2017mobilenets,sandler2018mobilenetv2,howard2019searching}, and the recent vision transformers (ViTs)~\cite{dosovitskiy2020vit, liu_2021_swin,liu2022swin}.

%% file: figs/overview.tex
\begin{figure*}[t]
    \small
    \centering
    \includegraphics[width=1.0\linewidth]{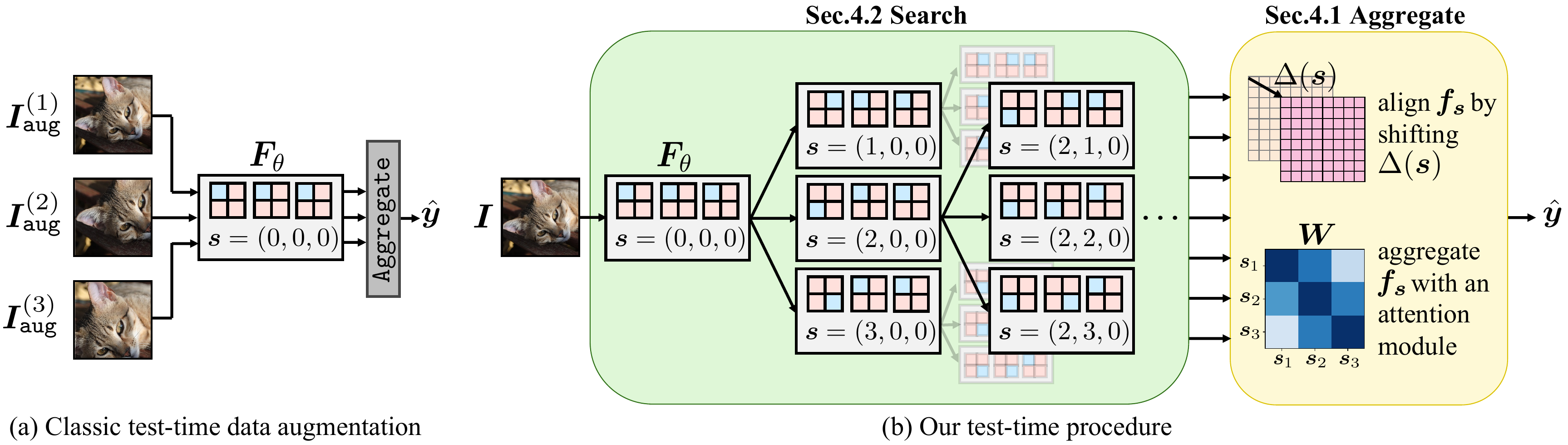}
    \vspace{-0.4cm}
    \caption{\textbf{Comparisons on test-time procedures.} {\bf (a)} In classic test-time augmentation, the output $\hat \vy$ is aggregated from different augmented images $\mI_{\texttt{aug}}$ feeding into the same model $\mF_\theta$ with default selection indices $\vs = (0,0,0)$. {\bf (b)} In our procedure, $\hat \vy$ is aggregated over one single image $\mI$ feeding into $\mF_\theta$ but activations are  extracted over a set of selection indices $\vs$. We apply a searching algorithm to search for the top-$B_{\texttt{ours}}$ selection indices $\vs$ based on a scoring function (Sec~\ref{sec:search}). We then aggregate (Sec~\ref{sec:aggregate}) the resulting feature set $\gF = \{\vf_\vs\}$ by first aligning each feature according to $\vs$ and then merging them using an attention aggregation module.
    }
    \vspace{-0.4cm}
    \label{fig:overview}
\end{figure*}

%% file: figs/subsampling.tex
    \centering
    \hspace{-0.5cm}
    \includegraphics[height=1.47cm]{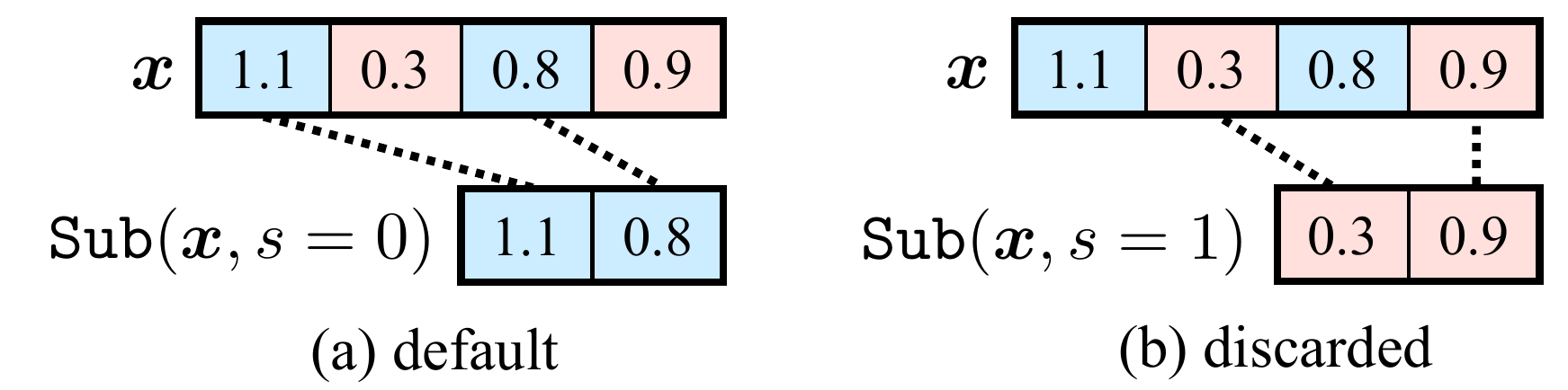}
    \vspace{-0.4cm}
    \captionof{figure}{\textbf{Subsampling by two.}}
    \label{fig:subsampling}

%% file: sec03_app.tex
\myparagraph{Classification formulation.} 
An image classification model with $K$ classes is trained over a dataset $\gD=\{(\mI, \vy)\}$ by minimizing the negative log-likelihood: 
\bea
\gL(\gD) = -\sum_{(\mI,\vy)\in\gD} \sum_{k=1}^K \vy[k] \log(\hat{\vy}[k](\mI)),
\eea
where, $\vy$ denotes the class label in one-hot representation, and $\hat{\vy}_c(\mI)$ denotes the predicted probability of class $c$. The deep net making the prediction can be defined as a composition of a feature extractor $F_\theta$ and a classifier $C_\phi$, \ie, 
\bea
\hat{\vy}(\mI) = C_\phi \circ F_\theta(\mI, \vs), \text{with } \vs=\mathbf{0}.
\eea
Consider a feature extractor $F_\theta$ consisting of $L$ subsampling layers, each with a subsampling factor $R^{(l)}$. We introduce a selection vector $\vs \in \gS$ to denote indices of the activations, \ie, a tuple of $s$ in~\equref{eq:sub} for each layer. The set 
$
\gS = \prod_{l=1}^L \{0,\dots,R^{(l)}\}
$ 
denotes the selection index for all the possible activations that can be extracted from a given image. The default forward pass corresponds to using $\vs=(0,0,0,...)=\mathbf{0}$ and all other $\gS\backslash\{\bf 0\}$ activations are disgarded.

\section{Our Approach}
Given a trained deep net with subsampling layers, we develop a test-time procedure that improves model performance. The approach is motivated by the observation that subsampling layers discard activations, \eg, a subsampling factor of two on a 2D feature map leads to losing $\frac{3}{4}$ of the spatial activations. As these discarded activations contain information about the input image, we believe they can be incorporated at test-time to improve model performance.

The idea is to keep a subset of the discarded activations and aggregate them into an improved prediction. 
To do so, we need to answer the following:
\begin{enumerate}[topsep=2pt]
    \item[(a)] {\bf How to better aggregate the activations?}
    Using all test-time transformations is not always a good idea~\cite{lyzhov2020greedy, shanmugam2021better}. Similarly, simply averaging all discarded activations leads to degraded performance.

    \item[(b)] {\bf Which of the discarded activations to retain?}
    Naively keeping all the discarded activations leads to an exponential growth in the number of feature maps which is impractical.
\end{enumerate}
To address (a), in~\secref{sec:aggregate}, we describe how we learn an aggregation function based on the attention mechanism. 
To address (b), in~\secref{sec:search}, we discuss how the learned attention values can be used as a search criterion for finding useful discarded activations. An overview of the proposed aggregate and search framework is visually illustrated in~\figref{fig:overview}(b).

\subsection{Aggregating selected activations for prediction} 
\label{sec:aggregate}
Given a set of indices $\hat{\gS}$, an image feature $\vf_s \in \sR^d$ is extracted for each $\vs \in \hat \gS$ to form a feature set 
\bea
\gF =  \{ \vf_\vs \mid \vs \in \hat \gS \}, \text{where } \vf_\vs = F_{\theta}(\mI; \vs)
\label{eq:feature_set}
\eea 
and $F_\theta(\mI; \vs)$ denotes the deep net's backbone.
The size of indices set $|\hat \gS|$ is equal to a user-specified test-time budget $B_{\texttt{Ours}}$ which corresponds to the number of forward passes needed at test time. These features are combined via an aggregation function $A: \sR^{d \times B_{\texttt{Ours}}} \mapsto \sR^{d}$ and passed to the pre-trained classifier $C_\phi$ to make a  prediction 
$\hat{\vy} = C_\phi \left( A (\gF) \right).
$

This aggregation function can be learned, \eg, following the setup of~\citet{shanmugam2021better} or it can be learning-free, \eg, following the classic TTA where the aggregation $A$ is simply the average of all $\vf_\vs$, \ie, $A(\gF) = \frac{1}{B_{\texttt{ours}}} \sum_{\vs \in \hat \gS} \vf_{\vs}$. 
We now discuss our proposed learned aggregation function based on attention, and a learning-free aggregation based on entropy. 

{\bf Learning an aggregation function.}
We propose a learnable multi-head attention module to be the aggregation function for two reasons: \textbf{(a)} The importance of one feature $\vf_\vs$ is relative to other features. \textbf{(b)} An attention module is a set operator~\cite{lee2019set} that can take an input of variable size for the feature set $\gF$. In other words, the aggregation function can be trained on one fixed budget and evaluated at \textit{any arbitrary testing budget}. Contrarily, the learned aggregation function proposed by~\citet{shanmugam2021better} is retrained for each test-time budget.

For each $\vf_\vs$, we learn its query $\vq_\vs$, key $\vk_\vs$, and value vector $\vv_\vs$ using a fully connected layer. The attention matrix $\mW$ is obtained by computing the inner product and normalizing through a softmax among the queries and keys for all $(\vs, \vs')$ pair,~\ie,
\bea
W_{\vs \vs'} = %
\exp(\vq_{\vs}^{\intercal} \vk_{\vs'}) / \left(
\sum_{\vs'' \in \hat \gS} \exp(\vq_{\vs}^{\intercal} \vk_{\vs''})\right).
\label{eq:attention-w}
\eea
The attention module's output is then used to learn an offset from the average features,~\ie,
\bea
A_{\tt learned}(\gF) = \frac{1}{B_{\texttt{ours}}}  \sum_{\vs \in \hat \gS}
\left(
\vf_{\vs} + \texttt{MLP} (\sum_{\vs' \in \hat \gS} W_{\vs\vs'} \vv_{\vs'})\right),
\label{eq:aggregate-learn}
\eea 
to obtain the aggregated feature over the set $\hat{\gS}$. Here, $\texttt{MLP}$ denotes an MLP model. %

\myparagraph{Learning-free aggregation}.  Inspired by~\citet{chun2022cyclic}, we use the entropy of the logits to quantify the confidence of the model in its prediction and weight each activation accordingly. Lower entropy indicates the more confident the model is for a prediction and thus it should receive a higher weight. We propose to weight each $\vf_\vs$ 
\bea
w_\vs = \frac{1}{Z_\vs}\left(1 - \frac{H(C_\phi(\vf_\vs))}{\log K}\right),
\label{eq:aggregate-entropy-weight}
\eea where $H(\cdot)$ is the entropy function, $K$ is the number of classes, and $Z_{\vs}$ is the normalization term such that $\sum_\vs w_{\vs} = 1$.
Overall, our training-free aggregation is as follows,
\bea
A_{\tt entropy}(\gF) = \sum_{\vs \in \hat \gS} w_\vs \vf_{\vs}.
\label{eq:aggregate-entropy}
\eea 

\begin{figure}[t]
    \centering
    \includegraphics[width=0.5\linewidth]{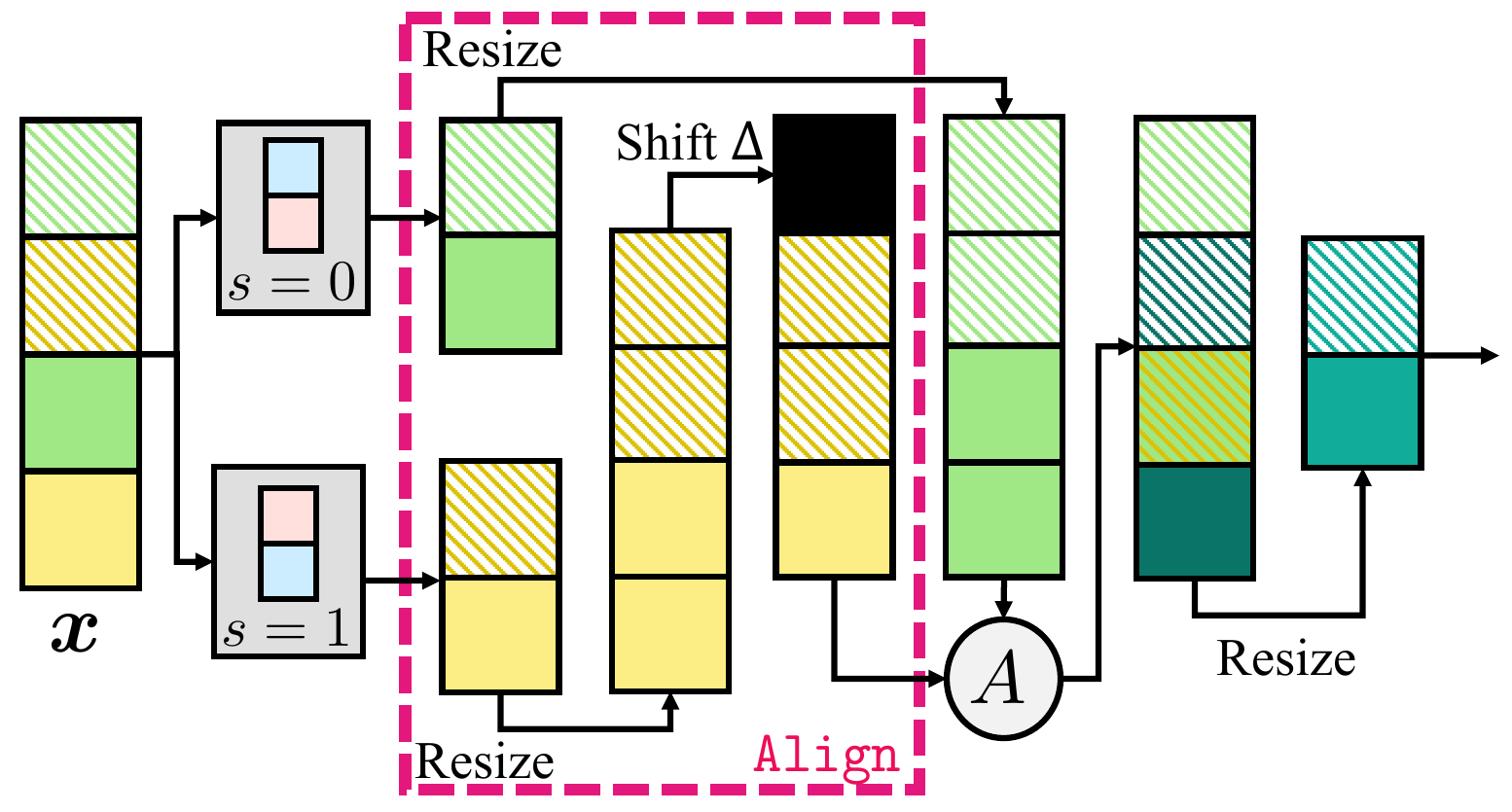}
    \vspace{-0.25cm}
    \caption{\textbf{Illustration of $\texttt{Align}_\vs$ with a single subsampling layer.}}
    \vspace{-0.45cm}
    \label{fig:align}
\end{figure}

{\bf\noindent Spatial alignment ($\texttt{Align}_\vs$).} 
To properly aggregate feature maps $\vf_\vs$ from different states $\vs$, we find it more beneficial to consider their shifting in feature space before aggregation to avoid spatial mismatch and adjust Eq.~\eqref{eq:feature_set} to
\bea
\gF =  \{ \vf_\vs \mid \vs \in \hat \gS \}, \text{where } \vf_\vs = \texttt{Align}_\vs(F_{\theta}(\mI; \vs)).
\label{eq:align-set}
\eea

We perform an alignment based on the relative shift with respect to the input resolution. $\texttt{Align}_\vs$ first resizes the activation maps to the input resolution, then shifts perform a shift $\Delta$ relative to the default activation $F_\theta(\mI; \vzero)$. In \figref{fig:align}, we illustrate this alignment using an input of length 4 with a single subsampling by 2. The {\color{OliveGreen} green activations} are extracted with $\vs=0$ and the {\color{myyellow} yellow activations} ($\vs=1$) are aligned by shifting with $\Delta=1$. 
For a specific $\vs$, we can compute the shift $\Delta$ as follows:
\bea
\nonumber
 \Delta =& \sum_{l=1}^{L} s_l \left(\prod_{l'=1}^{l} R^{(l')} \right) 
 \\
 =& s_1 +  s_2 R^{(1)} + s_3 R^{(1)} R^{(2)} + \cdots,
 \label{eq:offset}
\eea
where $\vs = (s_1,s_2,\dots,s_L)$ denotes the selection indices and $R^{(l)}$ denotes the subsampling rate at the $l^{\text{th}}$ layer.

\subsection{Searching for useful activations} 
\label{sec:search}
\begin{wrapfigure}[14]{r}{0.6\linewidth}
\vspace{-1.8cm}
\input{figures/fig_alg_alg}
\end{wrapfigure}

We have described how we aggregated a set of features $\hat\gS$. In this section, we will describe how to find this set of indices where the corresponding activations would benefit the model performance from the discarded set $\gS\backslash\{0\}$. As its size $|\gS|$ grows exponentially with the number of subsampling layers, naively iterating for all activations would be expensive. 
To address this, we propose a greedy search to gradually grow a set of selected activations within a given compute budget $B_{\texttt{Ours}}$. The high-level idea of the search algorithm is summarized in~\alggref{alg:search} and visualized in~\figref{fig:search}.\\
\indent{\bf Search procedure.} We formulate the task of finding a set of promising 
\begin{wrapfigure}[22]{r}{0.52\textwidth}
\vspace{-0.65cm}
\input{figures/fig_alg}
\end{wrapfigure}
  activation maps $\hat\gS$ as a search problem over the state space $\gS = \prod_{l=1}^L \{0,\dots,R^{(l)}-1\}$ where each state $\vs = (s_1,s_2,\dots,s_L)$ corresponds to a tuple of selection index for each of the subsampling layer $l$ in the pre-trained deep net. The \texttt{Search}, in~\alggref{alg:search}, returns all the states it has visited within a computation budget $B_{\tt ours}$. 

To implement this search, we utilize a priority queue $Q$ (lowest values are popped first) to determine which of the following states is more promising to visit next. The priority queue is sorted based on the values computed from a $\texttt{Criterion}(\vf_\vs)$ function.
In theory, this criterion can be any function that maps a feature map $\vf_\vs$ to a real number. Later in this section, we describe our proposed learned and learning-free criteria.

For each visited state $\vs$, we then add its ``neighboring states'' for a subsampling layer $l$. We define the neighbors of state $\vs=\mathbf{0}$ at the $0^{\text{th}}$ layer is the set $\cup_{i\in (1,2,\dots R^{(0)})} \{(i,0,0,\dots)\}$. The expanded layer $l$ is selected in a top-down fashion. Finally, we use a dictionary $E$ to keep track of the expanded layers for each state to avoid redundancies.

\myparagraph{Learned criterion.}
We prioritize expanding the node with the highest attention score from the attention $\mW$ in Eq.~\eqref{eq:attention-w} since the correspondent feature contributes the most in the final aggregated feature. Thus, we choose
\bea
\texttt{Criterion}_{\texttt learned}(\vf_{\vs}) = (\sum_{\vs' \in \hat \gS} W_{\vs \vs'})^{-1}.
\label{eq:criterion-learn}
\eea

\myparagraph{Learning-free criterion.} As in the aggregation, we found that entropy $H$ is a suitable choice for finding useful features, \ie, the search should emphasize on high-confidence regions of the feature space, therefore we choose
$ $
\bea %
\texttt{Criterion}_{\texttt entropy}(\vf_{\vs}) = \sum_{k=1}^K \hat{\vy}_\vs[k] \log \hat{\vy_\vs}[k]
= H(\hat{\vy_\vs}),
\label{eq:criterion-entropy}
\eea
where $\hat\vy_\vs = C_\phi(\vf_\vs)$ corresponds to the predicted probability from the classifier.

{\bf Extension to semantic segmentation.}
To extend the classification formulation to semantic segmentation, we view semantic segmentation as a per-pixel classification problem. For example, the entropy will then be computed for each pixel location. To aggregate the selected activations, we compute a per-pixel weight map for each activation instead of a scalar weight for the entire activation map.

%% file: figures/fig_alg_alg.tex
\begin{minipage}{\linewidth}
    \begin{algorithm}[H]
    \caption{\texttt{Search} for activations (Top-down)}\label{alg:search}
    \begin{algorithmic}[1]
    \State {\bf Inputs:} $\mI$, $\gS$, \texttt{Criterion}, Budget $B_{\texttt{Ours}}$
    \State $Q \leftarrow$ Init an empty priority queue
    \State {\color{NavyBlue}$E \leftarrow$ Init a dictionary of empty set $\forall \vs \in \gS$}
    \State {\color{OliveGreen}$\hat\gS = \{\}$} {\color{gray}\# Keeps track of returned states}
    \State $Q.\text{insert}((\textbf{0}, 0))$
    \While{$|\hat\gS| \leq B_{\texttt{Ours}}$}
        \State $\vs = Q.\text{pop}()$
        \State $\color{Mahogany}l = \min(\{1,\dots,L\} \backslash E[\vs])$ {\it \color{gray}\# Top-down}
        \State {\color{NavyBlue}$E[\vs].\text{add}(l)$ {\color{gray}\it \# Keeps track of expanded $l$}}
        \ForAll{$\vs' \in \text{Neighbors}(\vs,l)$}
        \State {\color{gray} \# $Q$ sorted by \texttt{Criterion}}
        \State $Q.\text{insert}((\vs', \texttt{Criterion}(\vs', \mI)))$
        \State {\color{OliveGreen}$\hat\gS.\text{add}(\vs')$}
        \EndFor
    \EndWhile
    \State \textbf{Return:} $\color{OliveGreen}\hat\gS$
    \end{algorithmic}
    \end{algorithm}
\end{minipage}

%% file: figures/fig_alg.tex
\centering
\includegraphics[width=\linewidth]{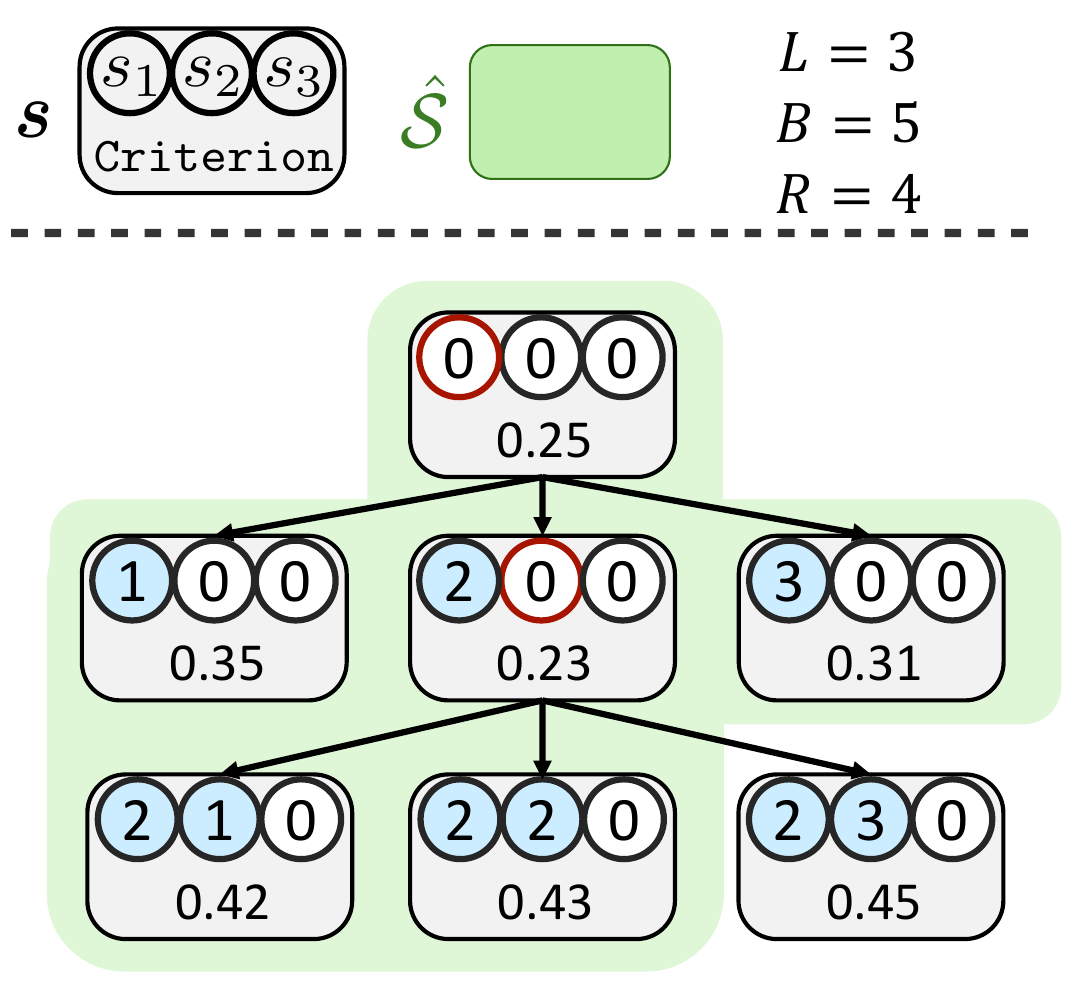}
\caption{\textbf{Search for activations.} From the initial state $(0, 0,0)$, we add its 3 neighbors ({\color{Mahogany}$l=1$}) in a top-down fashion. Next, state $(2, 0,0)$ has the lowest criterion, hence we further add its 3 neighbors ({\color{Mahogany}$l=2$}). Finally, the lowest-$B_{\texttt{ours}}$ states are returned in $\color{OliveGreen}\hat \gS$. We keep track of the expanded $\color{Mahogany}l$ in a {\color{NavyBlue}dictionary $E$}.
}
\label{fig:search}

%% file: sec04_exp.tex
\section{Experiments}

To validate the proposed test-time procedure, we conduct experiments on two computer vision tasks: image classification, and semantic segmentation. In the following, we provide the experiment setup and implementation details before discussing the results.

\input{sec04a_exp}

\input{sec04b_exp}

%% file: sec04a_exp.tex
\subsection{Image classification}
\label{sec:exp_imagenet}

\input{tables/tta-imagenet-expanded}

\myparagraph{Experiment setup.}
Recent TTA methods on image classification focus on learning TTA. Hence, we focus on comparison using the learned aggregation and criterion from our approach. The results of our learning-free version are included in the appendix.

We evaluate our learned methods on two image classification datasets, namely ImageNet~\cite{krizhevsky2012imagenet} and Flowers102~\cite{nilsback2008automated}. We provide experiments on additional datasets~\cite{krizhevsky2009learning} in the appendix.
For ImageNet, we use the pre-trained weights released by Pytorch. 
We randomly selected 20,000 and 5,000 images from \texttt{train} of ImageNet to be used for training and validation of the aggregation function. We report results on the \texttt{val} set of ImageNet.
For Flowers102, %
we fine-tune the pre-trained ImageNet weights from Pytorch on Flowers102 using the official training/validation/testing split for Flowers102. For both ImageNet and Flowers102, each image is first resized to 256px and cropped to 224px. 

Please note, this experiment setup differs from~\citet{shanmugam2021better} as their setup is non-conventional.~\citet{shanmugam2021better} resplit ImageNet's \texttt{val} set into train/validation/test sets. In other words, {\it their models are trained on a subset of ImageNet's original \texttt{val} set}. They also reported non-conventional ``multiple runs''. For each ``run'', they evaluate {\it the same trained model} over different \text{test subsets splits}, where each split is sampled (with replacement) from their ImageNet's test split\footnote{Please see their code~\url{github.com/divyashan/test-time-augmentation}) at $\tt{utils/evaluate.py}:L58$.}. Instead of following the unconventional setup, we conduct experiments using their released code on standard \texttt{val} split for each of the datasets. For multiple runs, each experiment is repeated 5 times with different random seeds, where the \textit{training split} is resampled, and the aggregation module is reinitialization. Overall, we observe the standard deviations for our method to be less than 0.05\%.

\input{tables/tta-flowers-expanded}

\myparagraph{Implementation details.}
Following~\citet{shanmugam2021better}, we choose ResNet18, ResNet50~\cite{he2016deep},   
 MobileNetV2~\cite{sandler2018mobilenetv2}, and InceptionV3~\cite{szegedy2016rethinking} to test the generalizability of our procedure on multiple backbones. Please refer to the supplementary for our definition of $F_\theta$ and $C_\phi$ and more results on additional backbones, \eg ResNext50~\cite{xie2017aggregated}, ShuffleNetV2~\cite{ma2018shufflenet}, Swin~\cite{liu_2021_swin}, and SwinV2~\cite{liu2022swin}.

To train our aggregation module $A$, we fix the budget $B_{\texttt{ours}}$ = 30 during training. To encourage the model to expand unseen nodes rather than focusing on seen ones, %
during training, we randomly sampled the nodes with probability inverse-correlated to its $\texttt{Criterion}$. We use the AdamW optimizer~\cite{loshchilov2017decoupled} and the cosine-annealing scheduler. The initial learning rate on ImageNet and Flowers102 is set to $1e^{-6}$.%
The training batch sizes are 32 for all datasets.

\myparagraph{Baselines.}
For baselines, we select several state-of-the-art TTA methods, namely GPS~\cite{lyzhov2020greedy}, AugTTA~\cite{shanmugam2021better}, and ClassTTA~\cite{shanmugam2021better}. For a fair comparison, each method is evaluated under the same test-time budget $B_\texttt{total}$, \ie, the number of total forward passes required to generate the final outputs.

Both baseline TTA methods and our procedure required their test-time budget. We denote the former one as $B_\texttt{tta}$ and the latter one as $B_\texttt{ours}$. Since the baseline TTA methods and ours are independent of each other, the overall budget $B_\texttt{total}$ is the product of $B_\texttt{tta}$ and $B_\texttt{ours}$.
For example, when comparing under the budget $B_\texttt{total}=150$ in~\tabref{table:tta-imagenet-expanded}, $(B_\texttt{tta}, B_\texttt{ours}) = (150, 1)$ is selected for the experiments w/o Ours, while $(B_\texttt{tta}, B_\texttt{ours}) = (10, 15)$ is selected for the experiments w/ Ours. Please refer to the supplementary for more detailed configuration.

We follow the \texttt{expanded} TTA policy settings~\cite{shanmugam2021better} and build a TTA pool containing various data transformations, such as \texttt{Flip}, \texttt{Colorization}, or \texttt{Blur}. 
The original \texttt{expanded} policy~\cite{shanmugam2021better} fixes $B_{\texttt{tta}}$ at 128. We increase the possible range of $B_{\texttt{tta}}$ for \texttt{expanded} policy to cover up to 1000. Please refer to supplementary for the detailed policy.

\myparagraph{Comparision to TTA.}
We report the comparison between the baseline TTA methods with and without ours in~\tabref{table:tta-imagenet-expanded} and~\ref{table:tta-flowers-expanded} on ImageNet and Flowers102 respectively. 

We report the performance under $B_\texttt{total} =$ 30, 100, and 150 in~\tabref{table:tta-imagenet-expanded}. We observe that the baseline TTA methods do not scale well when increasing $B_\texttt{total}$. The performance of GPS remains the same since the number of budget is fixed in their design; the gains on AugTTA are often minuscule or slightly negative; while ClassTTA usually suffers from using more budget due to its difficulty in converging. Our approach instead achieves consistent gain when using higher $B_\texttt{total}$. Moreover,  ours outperforms its counterpart in all cases.
On average, we improve the top-1 Acc. of TTA baselines by 0.87\%. Specifically, we improve GPS on average by 0.32\%, ClassTTA by 2.01\%, and AugTTA by 0.19\%.

In~\tabref{table:tta-flowers-expanded}, we report the performance under the same settings on Flowers102. The baseline TTA methods do not benefit much from using more $B_\texttt{total}$. Our approach outperforms its counterpart in most cases while remaining competitive in the rest. For example, %
we achieved the top-1 Acc. over all baselines by 0.57\%.

\begin{table*}[t!]
\begin{minipage}{0.3\linewidth}
    \input{tables/ablation_searchspace}

\end{minipage}
\hfill
\begin{minipage}{0.3\linewidth}
    \input{tables/ablation_merge}
\end{minipage}
\hfill
\begin{minipage}{0.3\linewidth}
    \input{tables/ablation_select}

\end{minipage}
\end{table*}
\myparagraph{Ablation study.}
We conduct ablation studies using ResNet-18 as the base architecture and report the performance of our method in our validation split of ImageNet without any TTA. We use $B_{\texttt{ours}}$=30 in these studies since the difference between different choices can be minuscule when the budget is small and the searching process is short.

\begin{figure}[t]
\begin{minipage}{0.54\linewidth}
    \input{figs/budget}
\end{minipage}
\hfill
\begin{minipage}{0.44\linewidth}
    \input{figs/learn_vs_acc}
\end{minipage}
\end{figure}

In~\tabref{table:ablation-searchspace}, we test other choices for searching the space $\gS$ in~\alggref{alg:search}: line 8. Specifically, we consider limiting the search space $\{1, \cdots, L\}$ by considering fewer layers. We ablate by removing the expansion on the first layer ($l=1$). We hypothesize that the layer has a small offset and therefore should have little impact on final outputs. On the other hand, we can omit the expansion on the last layer ($l=L$) because the large alignment necessary from the aggregation deteriorates the quality of the activations. Limiting the search space also greatly decreases the latency of our approach due to faster spatial alignment.

Overall, we observe that it is beneficial to limit the search space for better performance and faster speed. Overall, we find out that top-down search while omitting the last layer has the best performance in image classification. We choose to omit both the first and last layer as our default option because it balances between good performance and fast latency.

In~\tabref{table:ablation-agg}, we consider other aggregation methods besides our proposed one in~\equref{eq:aggregate-learn}. 
The baseline is to simply average all activations.
We report the performance of training-free aggregation by entropy-weighting in~\equref{eq:aggregate-entropy}.
Finally, we ablate the $\texttt{Align}$ in~\equref{eq:aggregate-learn} to show the importance of aligning feature maps before aggregation.
Our proposed learned aggregation yields the best performance. Additionally, without aligning feature maps, the performance drops significantly by 0.36\% in top-1 accuracy.

In~\tabref{table:ablation-criterion}, we test other choices for $\texttt{Criterion}$, determining which $\vs'$ to expand first in~\alggref{alg:search}. The baseline is to expand purely by random. One can suggest choosing $\vs'$ based on the offsets $\Delta$ (see~\equref{eq:offset}). We show that expanding the $\vs'$ with the lowest entropy (most confident prediction) is competitive when training is not feasible. Using learned attention $\mW$ from the aggregation module as the criterion yields the best performance.

\myparagraph{Analysis on budget $B_{\texttt{ours}}$.}
In~\figref{fig:budget}, we study the effect of the budget $B_{\texttt{ours}}$ to our procedure on the selected backbones.
We report the top-1 Acc. on our ImageNet testing split w/o any TTA methods.
Our method steadily gains improvements on its own when we increase the budget $B_{\texttt{ours}}$ from $1$ to roughly $10$. Performance gains are mostly saturated when the budget reaches $20$, \ie, more budgets yield limited gains. 

We provide additional results on various pre-trained backbones~\cite{rw2019timm} in the appendix, \eg MobileNetV3~\cite{howard2019searching} Multi-scale ViT (MViTv2)~\cite{li2022mvitv2, li2022mvitv2}, DenseNet~\cite{huang2017densely}, VGG~\cite{Simonyan15}, RepVGG~\cite{ding2021repvgg}, DeiT~\cite{touvron2021training}, CoaT~\cite{xu2021co}, ConvNeXTV2~\cite{liu2022convnet, woo2023convnext}, XCiT~\cite{ali2021xcit}, VOLO~\cite{yuan2022volo}, PvTV2~\cite{wang2021pyramid}, PvTV2~\cite{wang2022pvt}, Efficientformer~\cite{li2022efficientformer}, and ShuffleNet~\cite{zhang2018shufflenet}.

\myparagraph{Analysis of the proposed criteria.} 
We present a visualization in~\figref{fig:learned_vs_acc} demonstrating that our learned attention $\mW$ in~\equref{eq:aggregate-learn} is a good selection criterion. For the set of selection indices $\vs \in \gS$ over ImageNet (validation), we plot out the accuracy of its associated top-1 Acc. verses its learned criterion in~\equref{eq:criterion-learn} and the learning-free criterion based on entropy~\equref{eq:criterion-entropy}. That is, each point represents an experiment in which one sole $\vs \in \hat \gS$ is used for prediction and its associated top-1 Acc. Empirically, we observe that indices $\vs$ with lower $\texttt{Criterion}(\vf_\vs)$ tend to have better accuracy, hence it is reasonable to prioritize them when selecting $\hat \gS$.

%% file: tables/tta-imagenet-expanded.tex
\begin{table*}[t]
\captionof{table}{
    \textbf{Comparison to TTA methods on ImageNet with expanded~\cite{shanmugam2021better} TTA policies.} 
    We evaluate on ImageNet under $B_{\texttt{total}} \in \{30, 100, 150\}$ with various model architectures. For each $B_{\texttt{total}}$, we report the top-1 Acc. of baseline TTA methods with and without our learned approach. Whichever is the better one is bolded. The results of our learned procedure are highlighted. 
}
\label{table:tta-imagenet-expanded}
\centering
\setlength{\tabcolsep}{2pt}
\setlength{\tabcolsep}{2pt}
\resizebox{\linewidth}{!}{%
\begin{tabular}{ c c ccc c  ccc c ccc c ccc } 
\specialrule{.15em}{.05em}{.05em}

\multirow{2}{*}{TTA} & 
\multirow{2}{*}{Ours} & 
\multicolumn{3}{c}{ResNet18} && 
\multicolumn{3}{c}{ResNet50} && 
\multicolumn{3}{c}{MobileNetV2} && 
\multicolumn{3}{c}{InceptionV3} \\
& & 
$30$ & $100$ & $150$ && 
$30$ & $100$ & $150$ && 
$30$ & $100$ & $150$ && 
30 & 100 & 150 \\

\cmidrule(lr){1-17}
& \ding{55} & 
70.51 & 70.51 & 70.51 &&
76.50 & 76.50 & 76.50 &&
72.24 & 72.24 & 72.24 &&
71.48 & 71.48 & 71.48
\\
\rowcolor{LightCyan} \cellcolor{White}
\multirow{-2}{*}{GPS~\cite{lyzhov2020greedy}} 
& \ding{51} & 
\bf 70.74 & \bf 70.74 & \bf 70.69 &&
\bf 76.74 & \bf 76.84 & \bf 76.87 &&
\bf 72.37 & \bf 72.61 & \bf 72.58 &&
\bf 71.86 & \bf 72.05 & \bf 72.02
\\
\cmidrule(lr){1-17}
& \ding{55} & 
69.09 & 68.23 & 66.40 &&
75.40 & 74.88 & 73.56 &&
70.58 & 69.97 & 67.81 &&
70.80 & 70.39 & 70.34
\\
\rowcolor{LightCyan} \cellcolor{White}
\multirow{-2}{*}{ClassTTA~\cite{shanmugam2021better}} 
& \ding{51} & 
\bf 70.37 & \bf 70.36 & \bf 70.37 &&
\bf 76.58 & \bf 76.61 & \bf 76.65 &&
\bf 71.44 & \bf 71.68 & \bf 71.63 &&
\bf 71.93 & \bf 71.99 & \bf 72.00
\\

\cmidrule(lr){1-17}
& \ding{55} & 
70.55 & 70.66 & 70.28 &&
76.54 & 76.59 & 76.47 &&
72.33 & 72.42 & 72.46 &&
71.65 & 71.88 & 71.98 
\\
\rowcolor{LightCyan} \cellcolor{White}
\multirow{-2}{*}{AugTTA~\cite{shanmugam2021better}} 
& \ding{51} & 
\bf 70.75 & \bf 70.79 & \bf 70.74 &&
\bf 76.76 & \bf 76.84 & \bf 76.89 &&
\bf 72.41 & \bf 72.62 & \bf 72.58 &&
\bf 72.09 & \bf 72.24 & \bf 72.24
\\
\specialrule{.15em}{.05em}{.05em}
\end{tabular}}

\end{table*}

%% file: tables/tta-flowers-expanded.tex
\begin{table*}[t]
\caption{
    \textbf{Comparison to TTA methods on Flowers102 with expanded~\cite{shanmugam2021better} TTA policies.} 
    We evaluate on Flowers102 under $B_{\texttt{total}} \in \{30, 100, 150\}$ with various model architectures. For each $B_{\texttt{total}}$, we report the top-1 Acc. of baseline TTA methods with and without our learned approach. Whichever is the better one is bolded. The results of our learned procedure are highlighted. 
}
\label{table:tta-flowers-expanded}
\centering
\setlength{\tabcolsep}{2pt}

\setlength{\tabcolsep}{2pt}
\resizebox{\linewidth}{!}{%
\begin{tabular}{ c c ccc c ccc c ccc c ccc} 
\specialrule{.15em}{.05em}{.05em}

\multirow{2}{*}{TTA} & \multirow{2}{*}{Ours} & 
\multicolumn{3}{c}{ResNet18} && 
\multicolumn{3}{c}{ResNet50} && 
\multicolumn{3}{c}{MobileNetV2} && 
\multicolumn{3}{c}{InceptionV3}\\
& & 
$30$ & $100$ & $150$ && 
$30$ & $100$ & $150$ && 
$30$ & $100$ & $150$ && 
$30$ & $100$ & $150$ \\
\cmidrule(lr){1-17}
& \ding{55} & 
\bf 89.04 & 89.04 & 89.04 &&
\bf 91.12 & \bf 91.12 & 91.12 &&
89.85 & 89.85 & 89.85 &&
87.69 & 87.69 & 87.69
\\
\rowcolor{LightCyan} \cellcolor{White}
\multirow{-2}{*}{GPS~\cite{lyzhov2020greedy}} 
& \ding{51} & 
88.93 & \bf 89.20 & \bf 89.19 &&
91.05 & 91.02 & \bf 91.17 &&
\bf 89.90 & \bf 90.10 & \bf 90.05 &&
\bf 87.95 & \bf 87.93 & \bf 87.79

\\
\cmidrule(lr){1-17}
& \ding{55} & 
87.97 & 87.81 & 86.39 &&
90.84 & 90.19 & 90.06 &&
89.41 & 88.00 & 85.54 &&
87.10 & 87.41 & 84.81
\\
\rowcolor{LightCyan} \cellcolor{White}
\multirow{-2}{*}{ClassTTA~\cite{shanmugam2021better}} 
& \ding{51} & 
\bf 88.97 & \bf 89.15 & \bf 89.06 &&
\bf 91.04 & \bf 90.97 & \bf 91.02 &&
\bf 89.49 & \bf 89.58 & \bf 89.67 &&
\bf 87.43 & \bf 87.43 & \bf 87.46
\\

\cmidrule(lr){1-17}
& \ding{55} & 
88.73 & 88.86 & 88.55 &&
90.91 & 90.84 & 90.81 &&
\bf 90.10 & 90.00 & 89.82 &&
87.35 & 87.43 & 87.43
\\
\rowcolor{LightCyan} \cellcolor{White}
\multirow{-2}{*}{AugTTA~\cite{shanmugam2021better}} 
& \ding{51} & 
\bf 89.02 & \bf 89.55 & \bf 89.55 &&
\bf 91.14 & \bf 90.93 & \bf 90.97 &&
90.08 & \bf 90.13 & \bf 90.11 &&
\bf 87.62 & \bf 87.62 & \bf 87.79
\\
\specialrule{.15em}{.05em}{.05em}
\end{tabular}}

\end{table*}

%% file: tables/ablation_searchspace.tex
\captionof{table}{
    \textbf{Ablation on searching method.} 
    We report the top-1 Acc.~(\%) and latency~(img/s) on ImageNet with different choices of search space under $B_{\texttt{ours}}=30$.
    \label{table:ablation-searchspace}
}
\vspace{-0.4cm}
\centering
\scriptsize
\resizebox{1.05\linewidth}{!}{%
\begin{tabular}[t]{ l c c }
\specialrule{.15em}{.05em}{.05em}
{{\alggref{alg:search}: line 7}}
& {Acc.$\uparrow$} & {Latency$\downarrow$}
\\
\cmidrule(lr){1-3}
$\{ 1, \cdots, L\}$ & 79.64 & 53.46 \\ 
$\{ 1, \cdots, L-1\}$ & {79.52} & {25.06} \\
$\{ 2, \cdots, L\}$ & \textbf{80.06} & 25.84\\
$\{ 2, \cdots, L-1\}$ & \underline{79.88} & {21.21} \\
\specialrule{.15em}{.05em}{.05em}
\end{tabular}
}

%% file: tables/ablation_merge.tex
\captionof{table}{
    \textbf{Ablation on aggregation.} We report the top-1 Acc.~(\%) and latency~(img/s) on ImageNet with different choices of $A(\gF)$ under $B_{\texttt{ours}}=30$.
    \label{table:ablation-agg}
}
\vspace{-0.245cm}
\centering
\scriptsize
\resizebox{1.12\linewidth}{!}{%
\begin{tabular}{ l c c } 
\specialrule{.15em}{.05em}{.05em}
{$A(\gF)$} 
& {Acc.$\uparrow$} & {Latency$\downarrow$}
\\
\cmidrule(lr){1-3}
\texttt{Avg} & {79.38} & {19.55} \\
{\texttt{Entropy}} & {79.44} & {19.56} \\
Ours w/o $\texttt{Align}$ & \underline{79.52} & 20.09\\
Ours (w/ $\texttt{Align}$) & \bf 79.88 & 21.21 \\
\specialrule{.15em}{.05em}{.05em}
\end{tabular}
}

%% file: tables/ablation_select.tex
\captionof{table}{
    \textbf{Ablation on searching criterion.} We report the top-1 Acc.~(\%) and latecny~(img/sec) on ImageNet with different choices of $\texttt{Criterion}(\vf_\vs)$ under $B_{\texttt{ours}}=30$.
    \label{table:ablation-criterion}
}
\vspace{-0.4cm}
\centering
\scriptsize
\resizebox{1.1\linewidth}{!}{%
\begin{tabular}[t]{l c c  } 
\specialrule{.15em}{.05em}{.05em}
{$\texttt{Criterion}(\vf_\vs)$} & Acc.$\uparrow$ & Latency$\downarrow$ \\
\cmidrule(lr){1-3}
Random  & 
79.80 & 13.53
\\
$\Delta$ & 
79.72 & 12.65 
\\
{$H(\hat\vy)$ (Eq.~\eqref{eq:criterion-entropy})} & 
\underline{79.86} & 21.45
\\
{Ours (Eq.~\eqref{eq:criterion-learn})} & 
\bf{79.88} & 21.21
\\
\specialrule{.15em}{.05em}{.05em}
\end{tabular}
}

%% file: figs/budget.tex
    \centering
    \includegraphics[width=\linewidth]{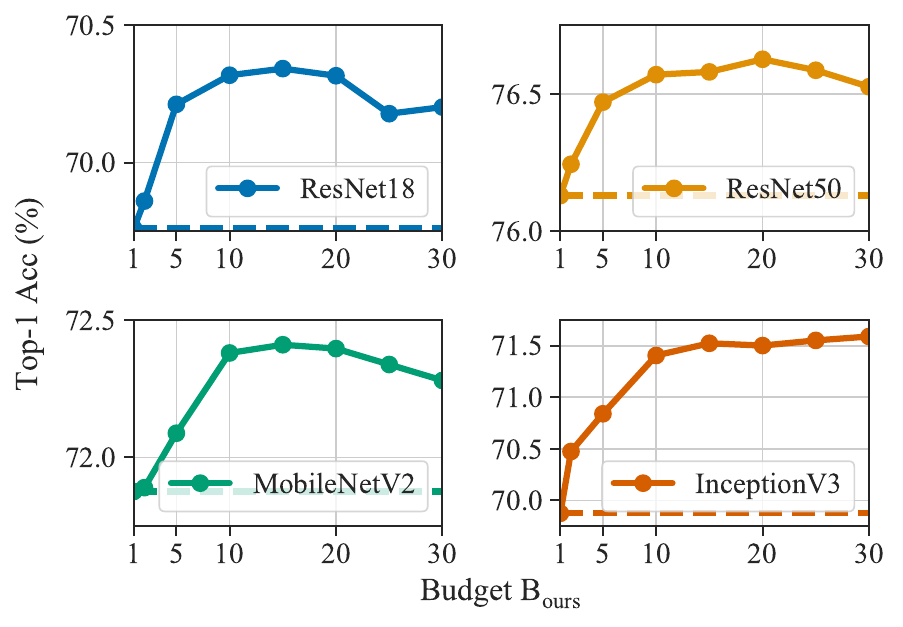}
    \caption{\textbf{Acc. vs. budget.}
    We observe an initial gain in Acc. when increasing the budget $B_\texttt{ours}$. The improvement plateaus when $B_\texttt{ours}$ reaches about 15.
    }
\label{fig:budget}

%% file: figs/learn_vs_acc.tex
\centering
\includegraphics[width=\linewidth]{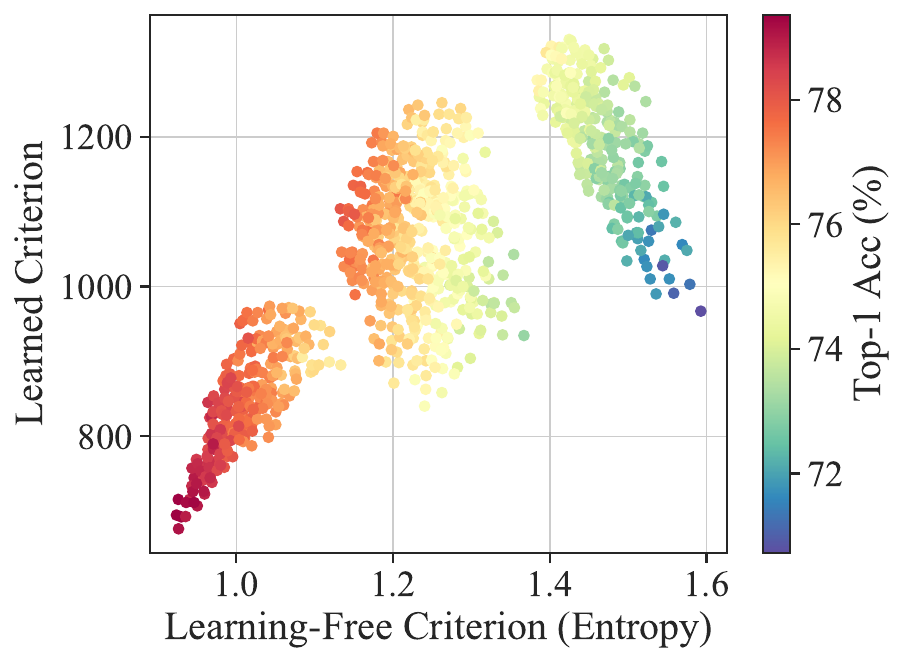}
\caption{
    \textbf{$\texttt{Criterion}_{\tt learned}(\vf_\vs)$ vs. $H(\hat \vy_\vs)$ vs. Accuracy.} Each point represents an experiment in which one sole $\vs \in \hat \gS$ is used for prediction and its associated top-1 Acc. over its learned criterion and entropy. 
}
\label{fig:learned_vs_acc}

%% file: sec04b_exp.tex
\subsection{Semantic segmentation}

\input{tables/ensemble-ade20k}
\myparagraph{Experiment setup.} 
As semantic segmentation has not been studied in learned TTA baselines~\cite{lyzhov2020greedy, shanmugam2021better}, we consider the non-learned TTA setting for this section. We conduct experiments using the non-learned version of our approach on CityScapes and ADE20K datasets~\cite{cordts2016Cityscapes, zhou2017scene, zhou2019semantic}.
We test on several semantic segmentation backbones and decoders, including ResNets~\cite{he2016deep}, MobileNets~\cite{howard2017mobilenets}, FCN~\cite{long2015fully}, DeepLab~\cite{chen2017rethinking, chen2018encoder}, and MiT+SegFormer~\cite{xie2021segformer}. %

\myparagraph{Implementation details.} 
Our implementation is based on MMSeg (OpenMMLab Segmentation)~\cite{mmseg2020}. It provides pre-trained weights and benchmarks on Cityscapes~\cite{cordts2016Cityscapes} and ADE20k~\cite{zhou2017scene, zhou2019semantic}. 
As for the searching strategy, different from the observation in image classification, we do not find it beneficial to limit the search space of~\alggref{alg:search}. All search spaces remain unchanged as $\{1, 2, \cdots, L\}$.
For evaluation metrics, we report the mIoU score on both datasets.

\myparagraph{Results.} In~\tabref{table:seg-main}, we report the results on Cityscapes and ADE20K semantic segmentation. We test with various combinations of encoder backbones and decoders for semantic segmentation to show that our method generalizes over architectures, including both conv. backbones~(MobileNet, ResNet-50) and transformer~(MiT) backbones.

We observe that our approach improves the performance over baseline models on all combinations of encoder backbones and decoders. The largest performance gains are 0.51 on Cityscapes mIoU with MiT + SegFormer ($B_{\texttt{ours}}=10$) and 0.27 on ADE20K mIoU with  MiT + SegFormer  ($B_{\texttt{ours}}=4$). These experiments show that our method can be extended to semantic segmentation with consistent improvements.

\myparagraph{Comparison to TTA.} In~\tabref{table:seg-main}, we also report the comparison to HorizotalFlip, which is a popular choice for TTA on semantic segmentation. Our test-time procedure further improves the performance when TTA is used. We observe a 0.30 gain in mIoU on MiT+Segformer ($B_{\texttt{ours}}=4$). The results show that our approach is orthogonal to TTA.

We provide additional results on various segmentation backbones in the appendix, \eg MobileNetV3+LRASPP~\cite{howard2019searching}, UNet~\cite{ronneberger2015u}, Swin~\cite{liu_2021_swin}, UperNet~\cite{xiao2018unified}, and Twins~\cite{chu2021twins}.

\subsection{Discussion \& limitation}
On both image classification and semantic segmentation, our proposed test-time procedure consistently demonstrates improvements over the baselines on a wide range of deep net architectures. We further point out that the achieved gains are considered significant in image classification and segmentation. However, the improved performance does come at a cost.

The limitation common among all test-time procedures is the increase in test-time computation. Multiple forward passes are needed to make a final prediction, and our method is no different. The increase in test-time computation means that the approach is not suitable for applications with real-time or low-power constraints. On the flip side, our method will be desirable for tasks where the accuracy is substantially more important than the compute time,~\eg, medical-related tasks. For these tasks, we believe our method is an appealing option to further improve performance.

%% file: tables/ensemble-ade20k.tex
\begin{table*}[t]
\centering
\caption{
    \textbf{Cityscapes and ADE20K results.} 
    We consider various encoder backbones and decoders w/ and w/o our procedure under different $B_{\texttt ours} \in \{4, 10\}$ and TTA (horizontal flip). 
    We report the mIoU score on both Cityscapes and ADE20K semantic segmentation. The results of our non-learned procedure are highlighted.
    Our procedure makes improvements on all combinations of architectures in both datasets. We also show that ours can improve the results of horizontal flip, proofing the complementary between our approach and standard TTA. 
}
\label{table:seg-main}
\setlength{\tabcolsep}{2pt}
\resizebox{\linewidth}{!}{%
\begin{tabular}{ c c ccc c ccc c ccc c ccc c ccc } 
\specialrule{.15em}{.05em}{.05em}

\multirow{2}{*}{Dataset} & \multirow{2}{*}{HFlip} & \multicolumn{3}{c}{ResNet50-FCN} && \multicolumn{3}{c}{ResNet50-DeepLab} && \multicolumn{3}{c}{MobileNet-FCN} && \multicolumn{3}{c}{MobileNet-DeepLab}
&& \multicolumn{3}{c}{MiT-SegFormer}
\\
& & 
\ding{55} & $4$ & $10$ && 
\ding{55} & $4$ & $10$ && 
\ding{55} & $4$ & $10$ && 
\ding{55} & $4$ & $10$ && 
\ding{55} & $4$ & $10$ \\
\cmidrule(lr){1-21}
\multirow{2}{*}{CityScape} 
& \ding{55} &
72.35 & \ours \underline{72.50} & \ours \bf{72.56} &&
79.60 & \ours \underline{79.72} & \ours \bf{79.73} &&
71.19 & \ours \underline{71.27} & \ours \bf{71.83} &&
75.32 & \ours \underline{75.58} & \ours \bf{75.60} &&
76.54 & \ours \underline{77.01} & \ours \bf{77.05}
\\
& \ding{51} &
72.71 & \ours \underline{72.88} & \ours \bf{72.97} &&
80.08 & \ours \bf{80.15} & \ours \underline{80.09} &&
71.69 & \ours \underline{72.03} & \ours \bf{72.10} &&
75.56 & \ours \underline{75.75} & \ours \bf{75.76} &&
76.84 & \ours \underline{77.11} & \ours \bf{77.12} 
\\
\cmidrule(lr){1-21}
\multirow{2}{*}{ADE20K} 
& \ding{55} &
35.94 & \ours \underline{36.12} & \ours \bf{36.20} &&
42.72 & \ours \underline{42.77} & \ours \bf{42.81} &&
19.55 & \ours \bf{19.57} & \ours \underline{19.56} &&
33.92 & \ours \underline{34.01} & \ours \bf{34.09} &&
37.41 & \ours \bf{37.68} & \ours \underline{37.67}
\\
& \ding{51} &
36.31 & \ours \underline{36.46} & \ours \bf{36.50} &&
43.02 & \ours \bf{43.07} & \ours \underline{43.06} &&
\underline{19.72} & \ours \underline{19.72} & \ours \bf{19.73} &&
34.18 & \ours \underline{34.20} & \ours \bf{34.28} &&
38.03 & \ours \bf{38.33} & \ours \underline{38.28}
\\
\specialrule{.15em}{.05em}{.05em}
\end{tabular}
}
\vspace{-0.2cm}
\end{table*}

%% file: sec05_conc.tex
\section{Conclusion}
We propose a test-time procedure that leverages the activation maps originally discarded by the subsampling layers. 
By solving a search problem to identify useful activations and then aggregating them together, via a learned aggregation module and criterion, the model can make more accurate predictions at test-time. On image classification and semantic segmentation tasks, we show that our approach is effective over nine different architectures. Additionally, it complements existing test-time augmentation approaches. These results suggest that our approach is a compelling method in addition to the existing testing TTA framework, especially for tasks that do not have real-time constraints.

%% file: sec06_appendix.tex
\setcounter{section}{0}
\renewcommand{\thesection}{A\arabic{section}}
\renewcommand{\thetable}{A\arabic{table}}
\setcounter{table}{0}
\setcounter{figure}{0}
\renewcommand{\thetable}{A\arabic{table}}
\renewcommand\thefigure{A\arabic{figure}}
\renewcommand{\theHtable}{A.Tab.\arabic{table}}%<---!!!!---
\renewcommand{\theHfigure}{A.Abb.\arabic{figure}}%<---!!!!---
\renewcommand\theequation{A\arabic{equation}}
\renewcommand{\theHequation}{A.Abb.\arabic{equation}}%<---!!!!---

{\noindent\bf \LARGE Appendix:}
\vspace{0.25cm}

\noindent The appendix is organized as follows:
\begin{itemize}[topsep=2pt]
\item In~\secref{sec:supp_exp_tta}, we provide additional details and comparisons to learned and non-learned TTA in other setups,~\eg, additional TTA policy or architectures.
\item In~\secref{sec:supp_exp_details}, we provide additional image classification results of our approach on CIFAR10 and CIFAR100 datasets~\cite{krizhevsky2009learning}. We also provide more detailed results on ImageNet for image classification, Cityscapes, and ADE20k for semantic segmentation.
\item In~\secref{sec:supp_impl_details}, we provide additional implementation details of our method, including how to implement subsampling layers for each of the corresponding backbones.
\end{itemize}

\section{Additional details and comparison to TTA}
\label{sec:supp_exp_tta}

\subsection{Details for integrating Ours with existing TTA methods.}
\label{sec:integration-tta}

Existing TTA methods take an input image and form an augmented image pool $\{ \mI_{{\tt aug}}^{(a)} \}_{a=1}^{B_{\tt tta}}$ to make a prediction by
\bea
    \hat \vy = T\left( \left\{\hat \vy^{(a)} \right\}_{a=1}^{B_{\tt tta}} \right), \text{where } \hat \vy^{(a)} = C_\phi(F_\theta(\mI_{{\tt aug}}^{(a)}; \vzero)),
\label{eq:tta-aggregate}
\eea where $T$ is an aggregation function (different from ours) merging all logits resulting from $B_{\tt tta}$ augmented images with default $\vs = \vzero$. To integrate Ours with existing TTA methods, we perform our search and aggregation method \textit{for each} augmented image to make the prediction $\hat\vy^{(a)}$ in~\equref{eq:tta-aggregate}, \ie,
\bea
    \hat{\vy}^{(a)} = C_\phi \left( A (\gF^{(a)}) \right)  \text{, where } \gF^{(a)} = \{ F(\mI_{{\tt aug}}^{(a)}; \vs) \mid \vs \in \hat \gS \}
\eea
following a simple aggregation $A$ as described in the main paper.
We use this described approach in our experiments when incorporating our approach with existing TTA methods.

\subsection{Learned TTA methods}

\myparagraph{A different TTA policy.} 
We conduct additional experiments following another TTA policy setting proposed by~\citet{shanmugam2021better},~\ie, the {\tt standard} TTA policy. The \texttt{standard} TTA consists of the following data transformations: \texttt{Flip}, \texttt{Scale}, and \texttt{FiveCrop}. 
The original \texttt{standard} policy fixes $B_{\texttt{tta}}$ at 30. We increase the possible range of $B_{\texttt{tta}}$ for \texttt{standard} policy to cover up to 190. Please refer to~\secref{sec:tta-policy} for the details.

\input{tables/tta-imagenet-standard}

\input{tables/tta-flowers-standard}
We report the comparison between
the baseline TTA methods with and without our learned approach in~\tabref{table:tta-imagenet-standard} and~\tabref{table:tta-flowers-standard} on ImageNet and Flowers102 respectively. We observe that, unlike our approach, the performance of TTA baselines is highly dependent on the TTA policy. Although the {\tt standard} policy improves the performance of TTA baselines, our approach shows a consistent gain over them, except for InceptionV3 with AugTTA.
As shown in~\tabref{table:tta-imagenet-standard}, on average, we improve the top-1 Acc. of TTA baselines by
0.9\% on ImageNet. Our approach improves GPS on average by 0.4\%,
ClassTTA by 2.5\%, and AugTTA by 0.1\%. 
As shown in~\tabref{table:tta-flowers-standard}, on average, we improve the top-1 Acc. of TTA baselines by
0.6\% on Flowers102.
Specifically, Our approach improves GPS on average by 0.1\%,
ClassTTA by 1.3\%, and AugTTA by 0.2\%.

\input{tables/tta-imagenet-expanded-app}
\myparagraph{More architectures.}
In~\tabref{table:tta-imagenet-expanded-app}, we report the comparison of our learned approach to baseline TTA methods on more architectures, including ResNext50~\cite{xie2017aggregated}, ShuffleNetV2~\cite{ma2018shufflenet}, Swin~\cite{liu_2021_swin}, and SwinV2~\cite{liu2022swin}. Specifically, we use these variants from torchvision~\cite{pytorch}: {\tt resnext50\_32x4d} for ResNext50, {\tt shufflenet\_v2\_x1\_0} for ShuffleNetV2, {\tt swin\_t} for Swin, and {\tt swin\_v2\_t} for SwinV2. We observe consistent gain across all the models on ImageNet.

\myparagraph{Computation budget.}
We compare the computation budget (MACs, and latency) on an image of resolution 224px by 224px between ours and baseline methods.
We measure the latency (ms/img) from end-to-end over 10 runs on an Nvidia A6000 GPU. We report the results of ResNet18 in~\tabref{table:computation_comparison} and MobileNetV2 in~\tabref{table:computation_comparison-mnv2}.
While GPS~\cite{lyzhov2020greedy} has less budget and latency, it does not achieve the best performance and does not scale with a higher budget as shown in the previous experiments. In terms of MACs, our approach requires more operations than ClassTTA~\cite{shanmugam2021better} and AugTTA~\cite{shanmugam2021better} on a lower budget. However, since our approach operates on downsampling layers only and shares most computation on non-downsampling layers, the overall latency is more or less similar to theirs.

\input{tables/computation_comparison}
\input{tables/computation_comparison-mnv2}

\subsection{Non-learned TTA methods}
% \ray{TODO: still does not make sense. If we use Eq.(9) for aggregation. what is the difference between TTA mean and Max? (For both Tab. A6 and A7, its very confusing. what is the difference.)}
For non-learned TTA methods, we follow~\citet{shanmugam2021better} and compared the {\tt MeanTTA} and {\tt MaxTTA} aggregation methods. {\tt MeanTTA} computes the average over all the augmented logits, while {\tt MaxTTA} picks the highest logit value for each class over all the augmented logits. We report the comparison of our non-learned approach to non-learned TTA methods,~\ie, we use ~\equref{eq:aggregate-entropy-weight} and \equref{eq:aggregate-entropy} for {\tt Aggregation} and \equref{eq:criterion-entropy} for {\tt Criterion}. As shown in~\tabref{table:tta-imagenet-expanded-non_learn}, we do not find it beneficial to use non-learned TTA methods with the expanded policy. We directly compare our non-learned approach without integrating TTA,~\ie, $B_{\tt tta}=1$, to non-learned TTA methods.
Our non-learned approach outperforms {\tt MeanTTA} and {\tt MaxTTA} without needing to increase the computation budget.

In~\tabref{table:tta-imagenet-standard-non_learn}, we report the non-learned TTA methods with or without our non-learned approach using the standard policy. We observe that although {\tt Mean} and {\tt MaxTTA} work on a smaller budget, they do not scale with higher budgets.
Our non-learned approach outperforms {\tt MeanTTA} under most budget settings while {\tt MaxTTA} is an ineffective non-learned TTA method.
\input{tables/tta-imagenet-expanded-non_learn}

\input{tables/tta-imagenet-standard-non_learn}

\subsection{Budget configuration}
With the integration approach described in~\secref{sec:integration-tta}, the total budget $B_{\tt total} = B_{\tt tta} \ times B_{\tt ours}$ is a product of the budget for existing TTA methods $B_{\tt tta}$ and ours $B_{\tt ours}$. We have the flexibility of balancing between the two to achieve the same total budget. Here, we document the choices used in our experiments.
\begin{itemize}
    \item For~\tabref{table:tta-imagenet-expanded}, \tabref{table:tta-flowers-expanded}, \tabref{table:tta-imagenet-standard}, \tabref{table:tta-flowers-standard}, \tabref{table:tta-imagenet-expanded-app}, \tabref{table:computation_comparison}, and~\tabref{table:computation_comparison-mnv2}, we use the following settings. When $B_{\tt total} = 30$, we use $(B_{\tt tta}, B_{\tt ours}) = (10, 3)$ with ours and $(B_{\tt tta}, B_{\tt ours}) = (30, 1)$ without ours.
When $B_{\tt total} = 100$, we use $(B_{\tt tta}, B_{\tt ours}) = (10, 10)$ with ours and $(B_{\tt tta}, B_{\tt ours}) = (100, 1)$ without ours.
When $B_{\tt total} = 150$, we use $(B_{\tt tta}, B_{\tt ours}) = (10, 15)$ with ours and $(B_{\tt tta}, B_{\tt ours}) = (150, 1)$ without ours.

\item For~\tabref{table:tta-imagenet-expanded-non_learn}, we use the following settings.
When $B_{\tt total} = 30$, we use $(B_{\tt tta}, B_{\tt ours}) = (1, 30)$ with ours and $(B_{\tt tta}, B_{\tt ours}) = (30, 1)$ without ours. When $B_{\tt total} = 75$, we only use $(B_{\tt tta}, B_{\tt ours}) = (75, 1)$ without ours.
When $B_{\tt total} = 150$, we  only use $(B_{\tt tta}, B_{\tt ours}) = (150, 1)$ without ours.

\item For~\tabref{table:tta-imagenet-standard-non_learn}, we use the following settings.
When $B_{\tt total} = 30$, we use $(B_{\tt tta}, B_{\tt ours}) = (15, 2)$ with ours and $(B_{\tt tta}, B_{\tt ours}) = (30, 1)$ without ours.
When $B_{\tt total} = 75$, we use $(B_{\tt tta}, B_{\tt ours}) = (15, 5)$ with ours and $(B_{\tt tta}, B_{\tt ours}) = (75, 1)$ without ours.
When $B_{\tt total} = 150$, we use $(B_{\tt tta}, B_{\tt ours}) = (15, 10)$ with ours and $(B_{\tt tta}, B_{\tt ours}) = (150, 1)$ without ours.
\end{itemize}

\section{Additional results}
\label{sec:supp_exp_details}
\subsection{CIFAR10 \& CIFAR100}
We use the publicly available implementation from PyTorch CIFAR Models (\url{https://github.com/chenyaofo/pytorch-cifar-models}), which supports a variety of pre-trained models on CIFAR10 and CIFAR100. Specifically, there are 19 models in total, including ResNet~\cite{he2016deep}, VGG~\cite{Simonyan15}, RepVGG~\cite{ding2021repvgg}, MobileNet~\cite{sandler2018mobilenetv2}, and ShuffleNet~\cite{zhang2018shufflenet} architectures. We report the experiments on all of them.

\begin{figure}[ht!]
    \centering
    \includegraphics[width=1\linewidth]{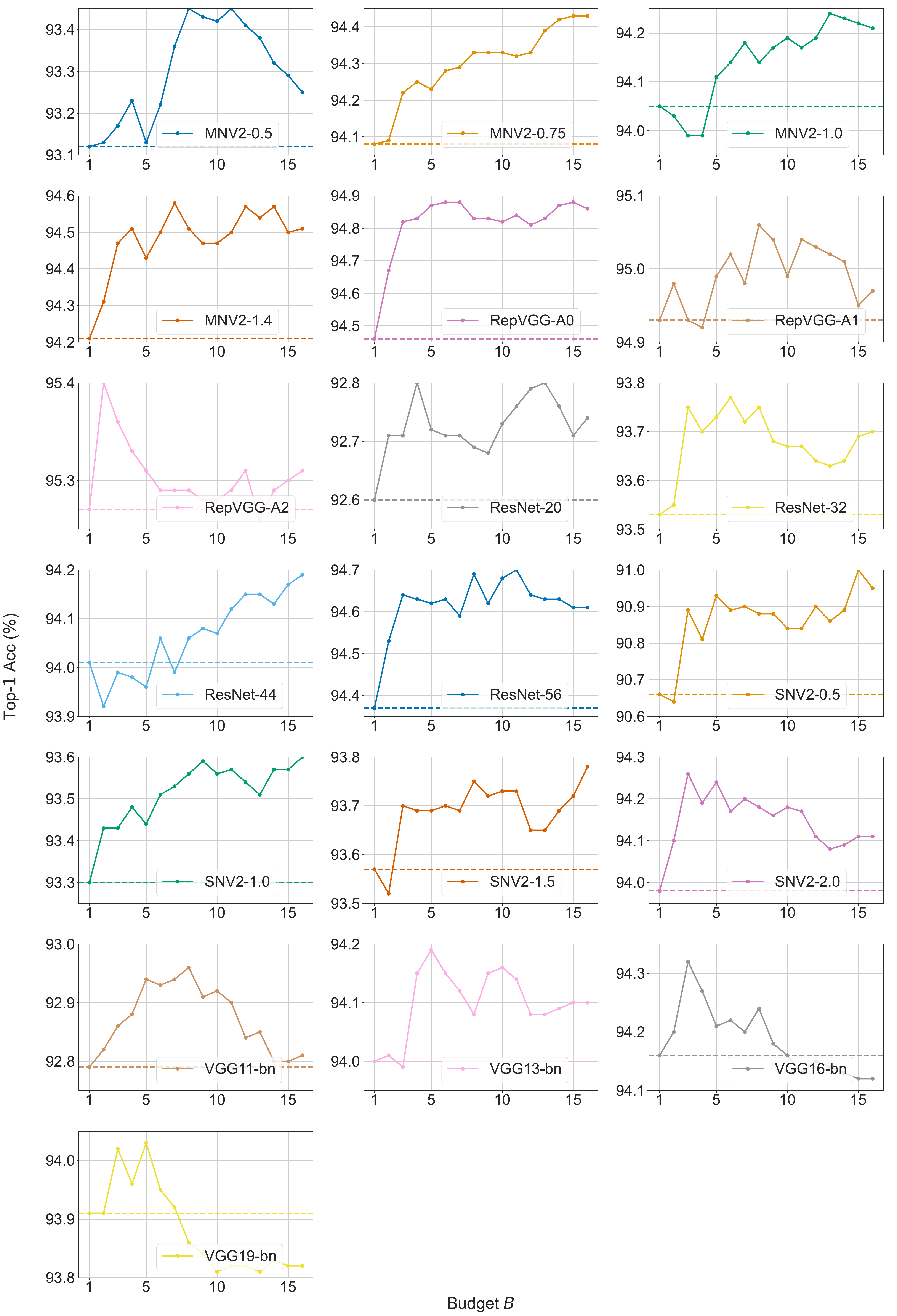}
    \vspace{-0.1cm}
    \caption{\textbf{Results on CIFAR10.} We report top-1 accuracy vs. budget $B_{\tt ours}$ used on CIFAR10 image classification. We use the following abbreviation in the figure legends: ``MNV2'' for MobileNetV2 and ``SNV2'' for ShuffleNetV2.
    }
    \label{fig:cifar10}
    x\vspace{-0.1cm}
\end{figure}

\begin{figure}[ht!]
    \centering
    \includegraphics[width=\linewidth]{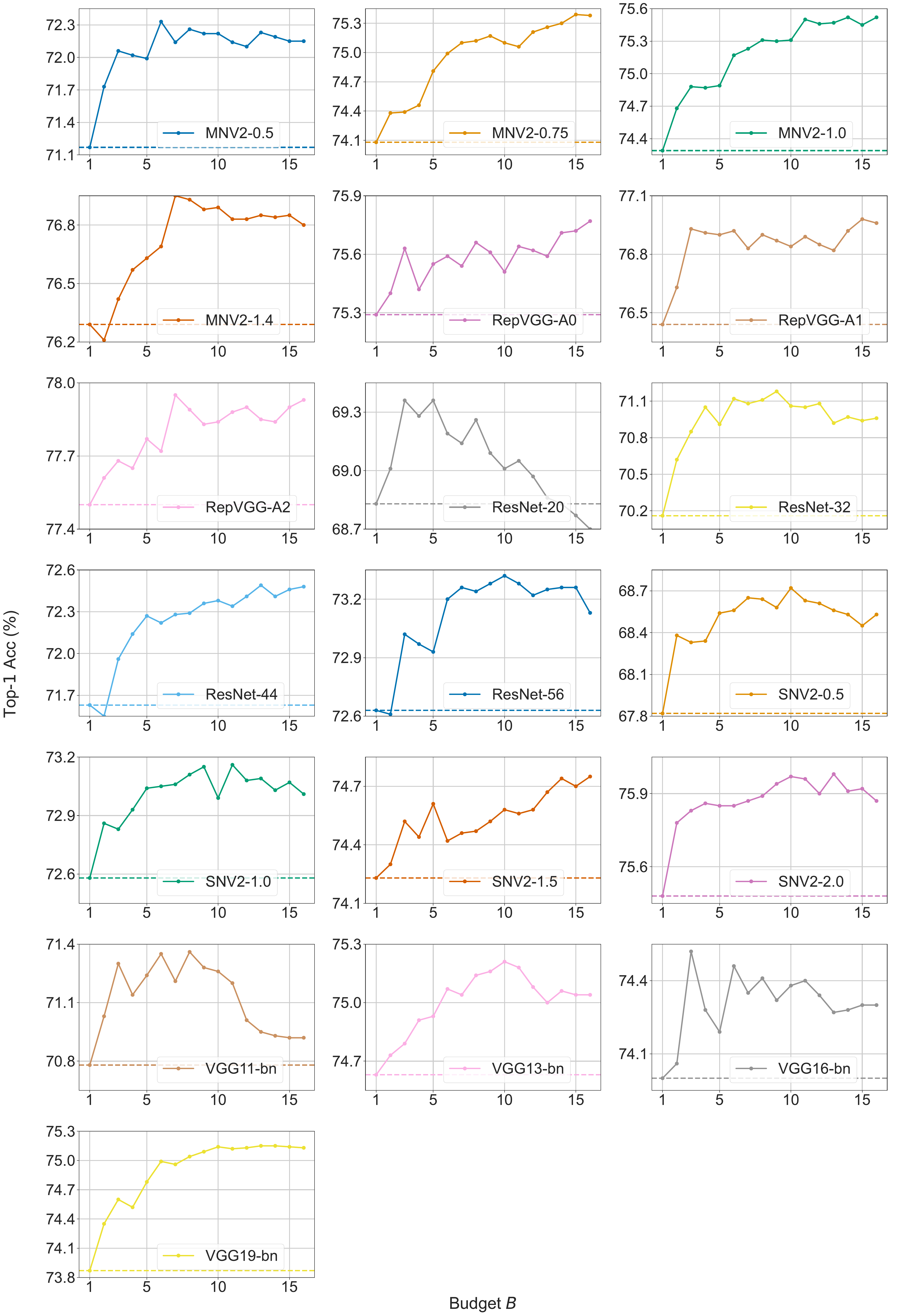}
    %\vspace{-0.1cm}
    \caption{\textbf{Results on CIFAR100.} We report top-1 accuracy vs. budget  $B_{\tt ours}$ used on CIFAR100 image classification. We use the following abbreviation in the figure legends: ``MNV2'' for MobileNetV2 and ``SNV2'' for ShuffleNetV2.
    }
    \label{fig:cifar100}
    %\vspace{-0.1cm}
\end{figure}

As in the main paper, we report the top-1 classification accuracy vs. the budget $B_{\tt ours}$ in~\figref{fig:cifar10} and~\figref{fig:cifar100} for CIFAR10 and CIFAR100 respectively. On both CIFAR10 and CIFAR100 datasets, we observe significant improvement in almost all scenarios except for VGG models. The performances of VGG models initially improves in low-budget setting and start to deteriorate quickly when given more budget. We suspect it is due to the fact that all subsampling layers in vanilla VGG are comprised of max-pooling layers. The local maximum remains very similar in discarded activation thus different state provides little additional information.

\begin{figure}[ht!]
    \centering
    \includegraphics[width=\linewidth]{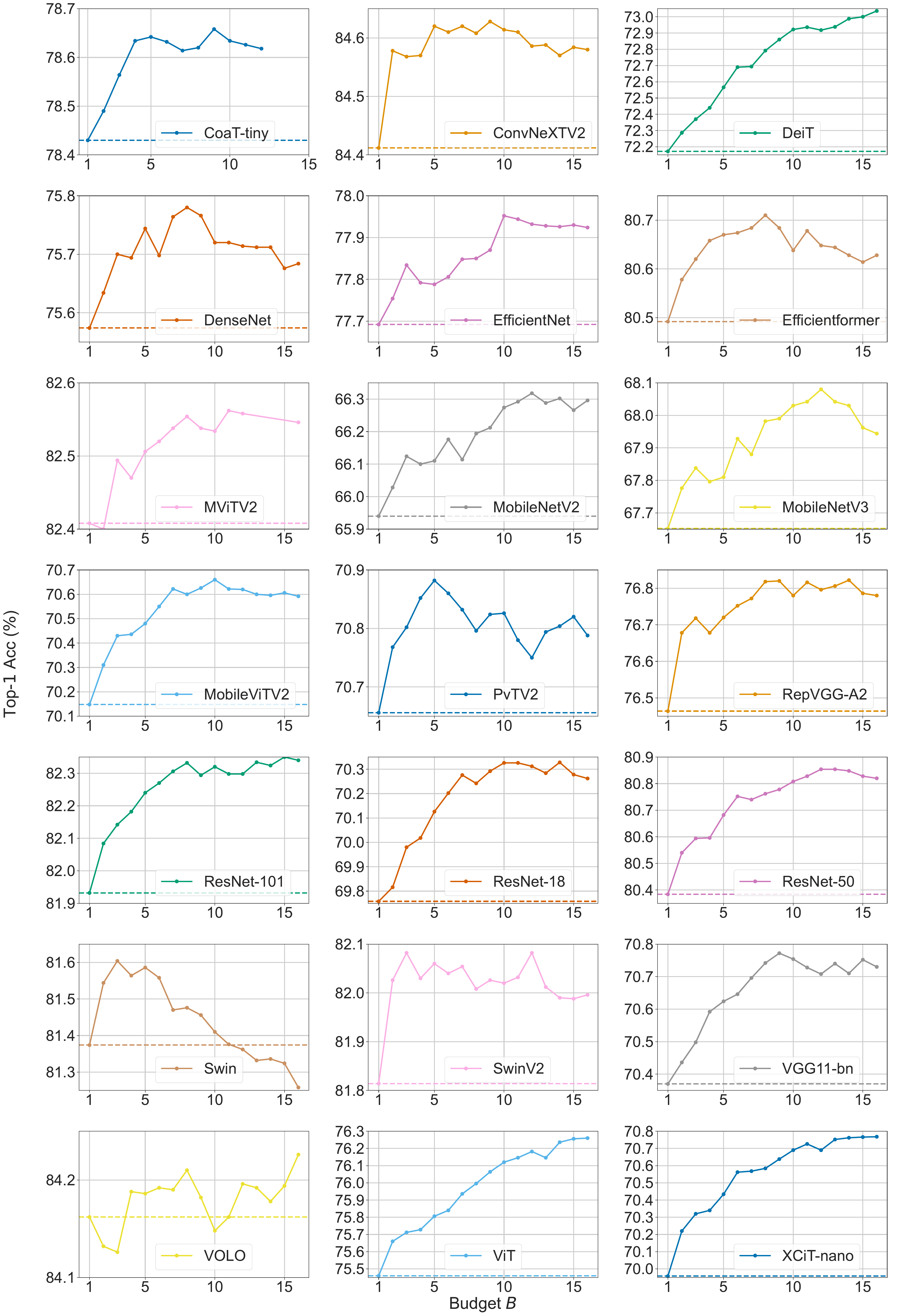}
    %\vspace{-0.1cm}
    \caption{\textbf{Additional results on ImageNet.} We report top-1 accuracy vs. budget $B_{\tt ours}$ used on ImageNet image classification.
    }
    \label{fig:more-imagenet}
    %\vspace{-0.1cm}
\end{figure}

\begin{figure}[ht!]
    \centering
    \includegraphics[width=\linewidth]{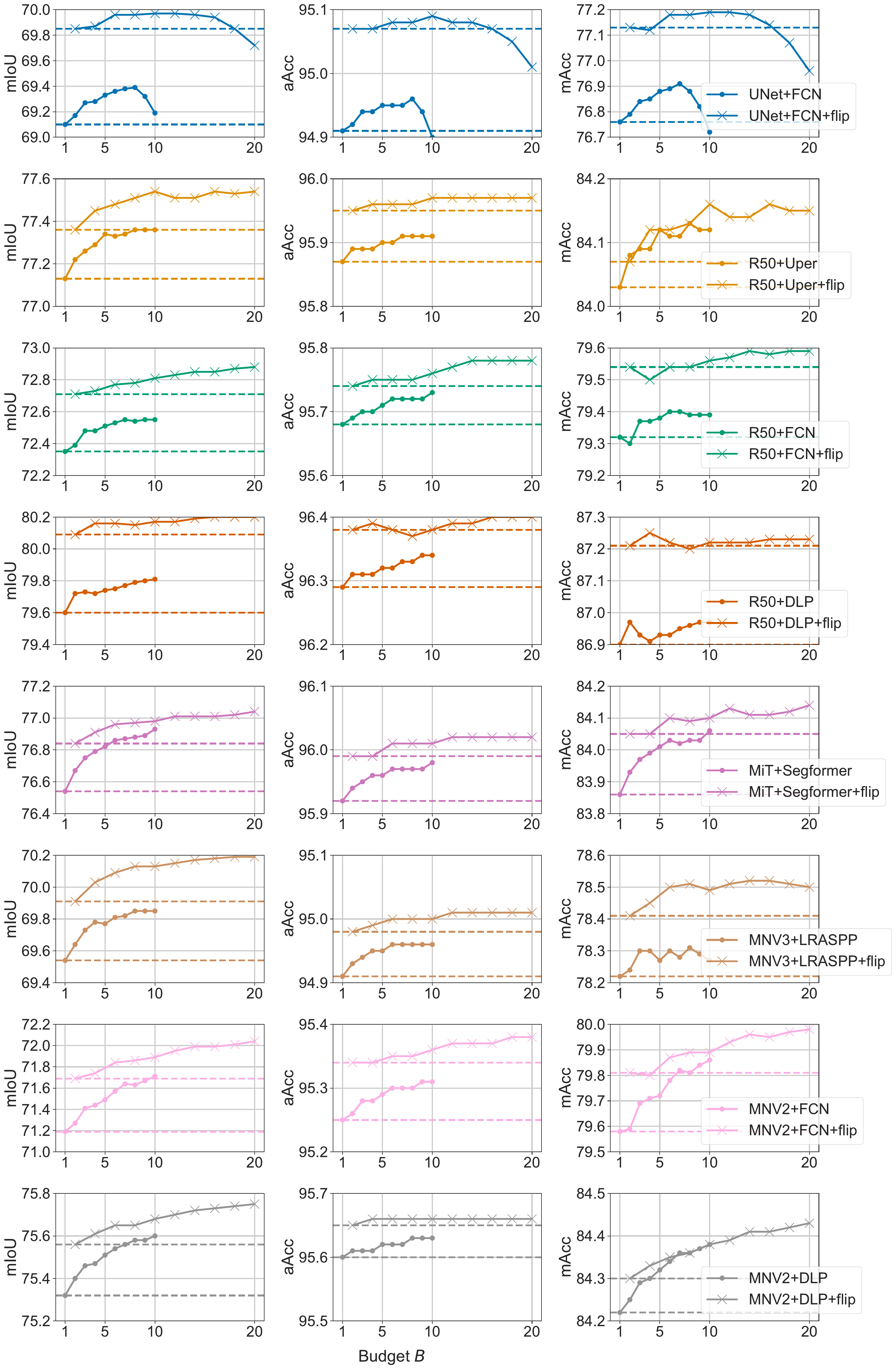}
    %\vspace{-0.1cm}
    \caption{\textbf{Additional results on Cityscapes.} We report mIoU/aAcc/mAcc vs. budget $B_{\tt total}$  used on Cityscapes semantic segmentation.
    We note that the $B_{\tt total}$  of ``model+flip'' is twice the $B_{\tt total}$  of ``model''.}
    \label{fig:more-cityscapes}
    %\vspace{-0.1cm}
\end{figure}

\begin{figure}[ht!]
    \centering
    \includegraphics[width=\linewidth]{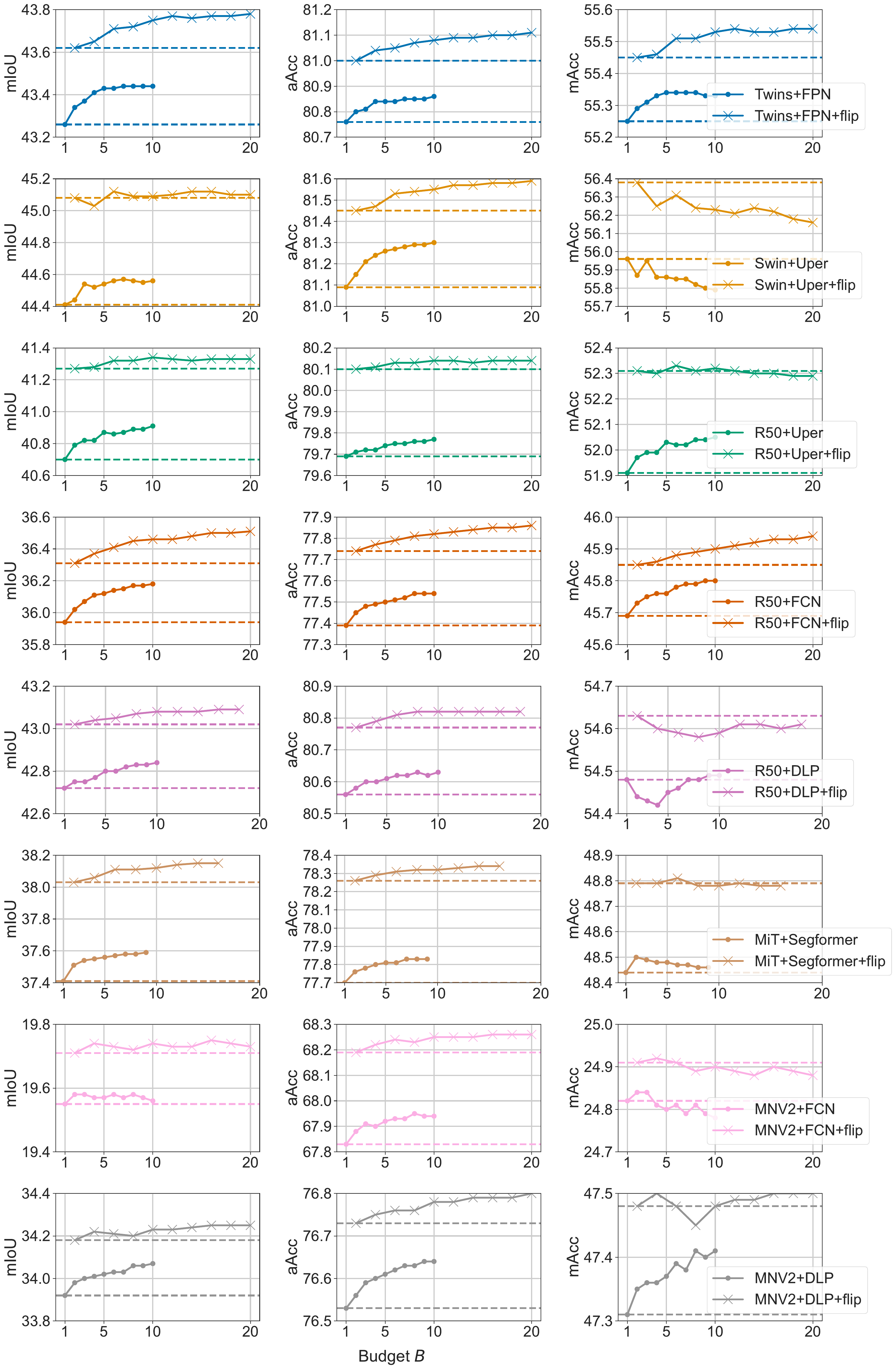}
    %\vspace{-0.1cm}
    \caption{\textbf{Additional results on ADE20K.} We report mIoU/aAcc/mAcc vs. budget  $B_{\tt total}$ used on ADE20K semantic segmentation. We note that the $B_{\tt total}$  of ``model+flip'' is twice the $B_{\tt total}$  of ``model''.
    }
    \label{fig:more-ade20k}
    %\vspace{-0.1cm}
\end{figure}

%\clearpage

%\subsection{CIFAR100}
\subsection{TIMM pretrained-weights on ImageNet}
In~\figref{fig:more-imagenet}, we provide additional results on TIMM~\cite{rw2019timm} backbones. These additional backbones include MobileNetV3~\cite{howard2019searching}, Multi-scale ViT (MViTv2)~\cite{li2022mvitv2}, DenseNet~\cite{huang2017densely}, VGG~\cite{Simonyan15}, RepVGG~\cite{ding2021repvgg}, DeiT~\cite{touvron2021training}, CoaT~\cite{xu2021co}, ConvNeXTV2~\cite{liu2022convnet, woo2023convnext}, XCiT~\cite{ali2021xcit}, VOLO~\cite{yuan2022volo}, PvTV2~\cite{wang2021pyramid, wang2022pvt}, Efficientformer~\cite{li2022efficientformer}. 
Specifically, we use these variants from TIMM: \texttt{mobilenetv3\_small\_100} for MobileNetV3, \texttt{mvitv2\_tiny} for MViTV2, \texttt{densenet121} for DenseNet, \texttt{vgg11\_bn} for VGG, \\
\texttt{repvgg\_a2} for RepVGG, \texttt{deit\_tiny\_patch16\_224} for DeiT, \texttt{volo\_d1\_224} for VOLO, \texttt{pvt\_v2\_b0} for PvTV2,
\texttt{coat\_tiny} for CoaT, \texttt{convnextv2\_tiny} for CoaT, and \texttt{efficientformer\_l1} for Efficientformer.

We observe that our approach consistently leads to better performance in most architectures. Our procedure excels in ViT-like architectures where subsampling layers with large subsampling rates $R$ are used.

\subsection{Cityscapes and ADE20K}
In~\figref{fig:more-cityscapes} and ~\figref{fig:more-ade20k}, we report 
additional results on Cityscapes and ADE20K semantic segmentation respectively. We plot out the performances on aAcc (mean accuracy of all pixel accuracy) and the mAcc (mean accuracy of each class accuracy) besides the mIoU score. We also provide additional results on more MMSeg~\cite{mmseg2020} backbones and decoders.
These additional architectures include MobileNetV3+LRASPP~\cite{howard2019searching}, UNet~\cite{ronneberger2015u}, Swin~\cite{liu_2021_swin}, UperNet~\cite{xiao2018unified}, and Twins~\cite{chu2021twins}. 
Specifically, we use these variants from MMSeg: \texttt{M-V3-D8} for MobileNetV3, \texttt{MiT-B0} for MiT, \texttt{Twins-PCPVT-S} for Twins, and \texttt{Swin-T} for Swin. 

As can be seen, our approach consistently leads to better performance, especially on mIoU and aAcc, in all architectures. Additionally, we show that our test-time procedure can further improve the performance when TTA (\texttt{horizontal-flip}) is used.

\subsection{Additional comparison to anti-aliasing downsampling}
To preserve the lost information from downsampling layers, a line of work, anti-aliased CNN~\cite{zhang2019making} inserts a {\tt max-blur-pool} layer to the pooling layer. Here, we demonstrate that our approach is orthogonal to the anti-aliasing method.

We report the performance of our approach on anti-aliased CNNs in~\tabref{table:antialias-cnn}. We conduct experiments using their pre-trained model and observe that applying our approach to anti-aliased CNNs also leads to consistent performance gain.

Finally, {\tt max-blur-pool} replaces max-pooling layers in CNN and requires training the whole model parameters; it is unclear how to apply it to ViTs which use patch merging instead of max-pooling. Our approach is generic and can be applied to both CNN and ViTs at test time while only training our additional attention module.

\input{tables/antialias-cnn}

%\clearpage
\section{Additional implementation details}
\label{sec:supp_impl_details}

\subsection{Attention module}
Recall~\equref{eq:attention-w} and~\equref{eq:aggregate-learn}, our attention module contains several trainable modules. Let $\vf_\vs = F_\theta(\mI; \vs) \in \sR^{h\times w\times c}$, then the query, key, and value features are defined as
\bea
\begin{cases}
    \vq_\vs &= w_q(\vf_\vs) \\
    \vk_\vs &= w_k(\vf_\vs) \\
    \vv_\vs &= \vf_\vs
\end{cases},
\eea
where $w_\vq$ and $w_\vk$ are both linear layers {\tt Linear(in=c, out=1)}. Finally, the $\texttt{MLP}$ in~\equref{eq:aggregate-learn} can be represented by a trainable tensor $w_o \in \sR^c$ so that
given a input $\vx \in \sR^c$,
\bea
    \texttt{MLP}(\vx) = w_o \odot \vx,
\eea where $\odot$ is the elementwise multiplication.
Overall, our attention module introduces three learnable parameters, i.e. $w_q, w_k$, and $w_o$, 

\subsection{TTA policy}
\label{sec:tta-policy}

\myparagraph{Standard policy.} The {\tt standard}~\cite{shanmugam2021better} TTA policy is composed of the following data transformations, \texttt{Flip}, \texttt{Scale}, and \texttt{FiveCrop}. 
The original \texttt{standard} policy fixes $B_{\texttt{tta}}$ at 30. We increase the possible range of $B_{\texttt{tta}}$ for \texttt{expanded} policy to cover up to 190. We detail the setting in~\tabref{table:tta-standard}.
For any budget $B_{\tt tta} < 190$, we sorted the data augmentations based on the intensity and sample the first $B_{\tt tta}$ transformations.

\input{tables/tta-standard}

\myparagraph{Expanded policy.} The {\tt expanded}~\cite{shanmugam2021better} TTA policy contains
various data transformations, such as \texttt{Saturation}, or \texttt{Blur}.
The original \texttt{expanded} policy fixes $B_{\texttt{tta}}$ at 128. We increase the possible range of $B_{\texttt{tta}}$ for \texttt{expanded} policy up to 8778. 
The pool contains two groups. Each data augmentation of the first group belongs to one of the 131 data augmentations listed in~\tabref{table:tta-expanded}. Each data augmentation of the second group is composed of two of the 131 data augmentations.
In total, our pool contains 8778 data augmentation.
For any budget $B_{\tt tta} < 8778$, we sorted the data augmentations based on the intensity and sample the first $B_{\tt tta}$ transformations.

\input{tables/tta-expanded}

\clearpage

\subsection{Computational settings}
All experiments are conducted on a single NVIDIA A6000 GPU.

\subsection{Modified subsampling layers}
\begin{figure}[h!]
    \centering
    \includegraphics[width=0.5\linewidth]{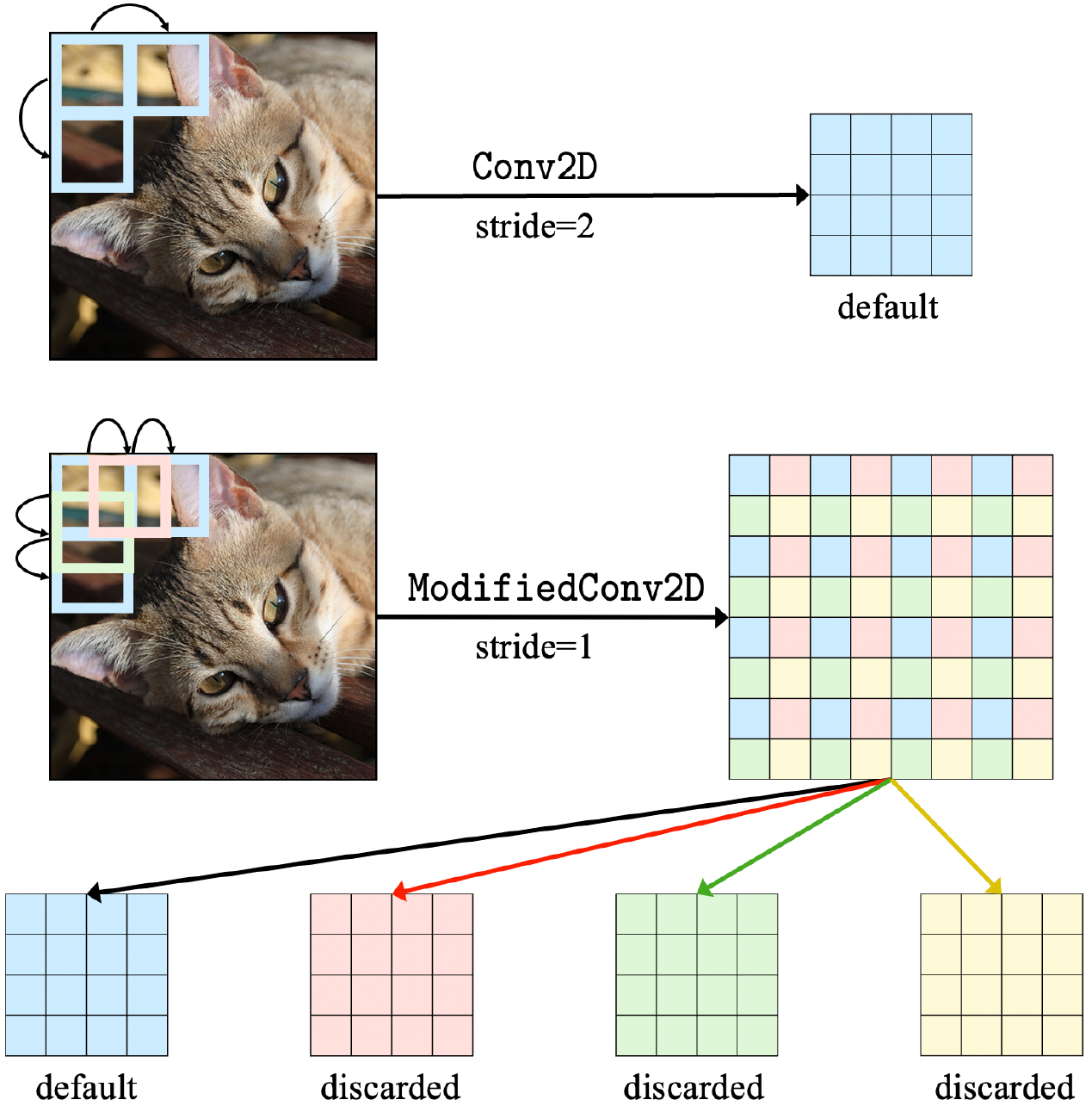}
    \caption{\textbf{Stride trick.} Instead of performing Conv2D with stride 2 four times, we can equivalently do a stride 1 convolution followed by a pixel unshuffle.
    }
    \label{fig:stride}
\end{figure}

In~\figref{fig:stride}, we illustrate an implementation trick for faster computation in~\alggref{alg:search}: line 10.
By changing the stride of the subsampling layer from their original stride value $R$ to $1$, we can get all the $R^2$ neighboring states (activations) in a single forward pass.

\subsection{Searching procedure}
For TIMM implementation, some models, eg. DenseNet and ViT, have fewer (less than 4) subsampling layers. We do not limit the searching space (\alggref{alg:search}: line 8) for them in our experiments. Some models, eg. PvTV2, have a larger subsampling rate $R$ ($=4$ or 16) on the first subsampling layer than the rest. We do not exclude the first layer from the searching space (\alggref{alg:search}: line 8) for them in our experiments.

\subsection{Spatial alignment}
For TIMM transformer-based backbones, the output of $F_\theta$ has the shape of \texttt{batch\_size} by \texttt{token\_length} by \texttt{channel\_size}. In order to perform our 2-dimensional spatial alignment, we reshape it to \texttt{batch\_size} by \texttt{sqrt}(\texttt{token\_length}) by \texttt{sqrt}(\texttt{token\_length}) by \texttt{channel\_size} if applicable. 

For example, the output shape of ViT, XCiT, CoaT, and DeiT features are not composite numbers so we do not reshape them. To aggregate the result from different states, we aggregate on the logits instead. 

Next, for both the upsampling and downsampling \texttt{Resize}  in~\figref{fig:align} in the spatial alignment, we use the nearest interpolation. 

\subsection{Image classification}
In this section, we illustrate how we modify an existing backbone using ResNet-18 as an example. %

For the Torchvision implementation of ResNet-18, we have listed the subsampling layers in~\tabref{table:sub-timm-r18}. In our implementation, we modify those layers to include them in the search space. Please refer to our code for more implementation details.

\begin{table}[t]
\centering
\small
\caption{\textbf{Subsampling layers of Torchvision Resnet-18.}}
\label{table:sub-timm-r18}
\begin{tabular}{ l l l c c c  c c c c c c c } 
\specialrule{.15em}{.05em}{.05em}
Id & Name & Type & $R$  \\
\midrule
1 & \texttt{conv1} & Conv2d & 2  \\
2 & \texttt{maxpool} & MaxPool2d & 2 \\
3 & \texttt{layer2/0} & BasicBlock & 2 \\
3-1 & \texttt{layer2/0/conv1} & Conv2d & 2 & \\
3-2 & \texttt{layer2/0/downsample/0} & Conv2d & 2\\
4 & \texttt{layer3/0} & BasicBlock & 2\\
4-1 & \texttt{layer3/0/conv1} & Conv2d & 2 \\
4-2 & \texttt{layer3/0/downsample/0} & Conv2d & 2 \\
5 & \texttt{layer4/0} & BasicBlock & 2\\
5-1 & \texttt{layer4/0/conv1} & Conv2d & 2 \\
5-2 & \texttt{layer4/0/downsample/0} & Conv2d & 2 \\

\specialrule{.15em}{.05em}{.05em}
\end{tabular}
\end{table}

%% file: tables/tta-imagenet-standard.tex
\begin{table*}[h!]
% \small
\captionof{table}{
    \textbf{Comparison to TTA methods on ImageNet with standard~\cite{shanmugam2021better} TTA policies.} 
    We evaluate on ImageNet under $B_{\texttt{total}} \in \{30, 100, 150\}$ with various model architectures. For each $B_{\texttt{total}}$, we report the top-1 Acc. of baseline TTA methods with and without our learned approach. Whichever is the better one is bolded. The results of our learned procedure are highlighted. 
}
\label{table:tta-imagenet-standard}
\centering
\setlength{\tabcolsep}{4pt}

\resizebox{\linewidth}{!}{%}
\begin{tabular}{ c c c ccc c ccc c ccc c ccc } 
\specialrule{.15em}{.05em}{.05em}

\multirow{2}{*}{TTA} & \multirow{2}{*}{Ours} & 
\multicolumn{3}{c}{ResNet18} && 
\multicolumn{3}{c}{ResNet50} && 
\multicolumn{3}{c}{MobileNetV2} && 
\multicolumn{3}{c}{InceptionV3}\\
& & 
$30$ & $100$ & $150$ && 
$30$ & $100$ & $150$ && 
$30$ & $100$ & $150$ && 
$30$ & $100$ & $150$ 
\\
% \cmidrule(lr){1-14}
% \multirow{6}{*}{\rotatebox[origin=c]{90}{ResNet18}} 
% \multirow{2}{*}{Mean} 
% & w/o & 
% 67.82 & 65.33 & & 
% & 80.66 & & 
% 64.96 & 63.72 \\
% % & w/ & x & 72.34 \\
% & w/ & 
% 67.55 & 67.48 & & & & & 65.01 & 64.77 
% \cmidrule(lr){1-14}
% \multirow{2}{*}{Cyclic} 
% \\
% \\
\cmidrule(lr){1-17} 
& \ding{55} & 
70.38 & 70.38 & 70.38 &&
76.37 & 76.37 & 76.37 &&
72.27 & 72.27 & 72.27 &&
71.74 & 71.74 & 71.74
\\
\rowcolor{LightCyan} \cellcolor{White}
\multirow{-2}{*}{GPS~\cite{lyzhov2020greedy}}
& \ding{51} &
\bf{70.60} & \bf{70.72} & \bf{70.69} &&
\bf{76.67} & \bf{76.81} & \bf{76.84} &&
\bf{72.38} & \bf{72.64} & \bf{72.59} &&
\bf{72.13} & \bf{72.29} & \bf{72.30}
\\

\cmidrule(lr){1-17}
& \ding{55} & 
70.06 & 67.77 & 65.46 &&
76.26 & 74.55 & 72.65 &&
71.79 & 69.19 & 66.74 &&
71.77 & 70.54 & 70.07
\\
\rowcolor{LightCyan} \cellcolor{White}
\multirow{-2}{*}{ClassTTA~\cite{shanmugam2021better}} 
& \ding{51} & 
\bf 70.70 & \bf 70.73 & \bf 70.69 &&
\bf 76.76 & \bf 76.83 & \bf 76.85 &&
\bf 72.32 & \bf 72.34 & \bf 72.29 &&
\bf 72.26 & \bf 72.43 & \bf 72.41
\\

\cmidrule(lr){1-17}
& \ding{55} & 
70.78 & 70.83 & 70.82 &&
76.68 & 76.67 & 76.66 &&
72.59 & 72.59 & \bf 72.61 &&
72.22 & \bf 72.99 & \bf 73.02
\\
\rowcolor{LightCyan} \cellcolor{White}
\multirow{-2}{*}{AugTTA~\cite{shanmugam2021better}} 
& \ding{51} & 
\bf 70.89 & \bf 70.85 & \bf 70.86 &&
\bf 76.74 & \bf 76.82 & \bf 76.86 &&
\bf 72.60 & \bf 72.69 & 72.58 &&
\bf 72.42 & 72.51 & 72.46
\\
\specialrule{.15em}{.05em}{.05em}
\end{tabular}
}

\end{table*}

%% file: tables/tta-flowers-standard.tex
\begin{table*}[h!]
% \small
\centering
\setlength{\tabcolsep}{4pt}
\captionof{table}{
    \textbf{Comparison to TTA methods on Flowers102 with standard~\cite{shanmugam2021better} TTA policies.} 
    We evaluate on Flowers102 under $B_{\texttt{total}} \in \{30, 100, 150\}$ with various model architectures. For each $B_{\texttt{total}}$, we report the top-1 Acc. of baseline TTA methods with and without our learned approach. Whichever is the better one is bolded. The results of our learned procedure are highlighted. 
}
\label{table:tta-flowers-standard}
\resizebox{\linewidth}{!}{%}
\begin{tabular}{ c c ccc c ccc c ccc c ccc} 
\specialrule{.15em}{.05em}{.05em}

\multirow{2}{*}{TTA} & \multirow{2}{*}{Ours} & 
\multicolumn{3}{c}{ResNet18} && 
\multicolumn{3}{c}{ResNet50} && 
\multicolumn{3}{c}{MobileNetV2} && 
\multicolumn{3}{c}{InceptionV3}\\
& & 
$30$ & $100$ & $150$ && 
$30$ & $100$ & $150$ && 
$30$ & $100$ & $150$ && 
$30$ & $100$ & $150$ \\
% \cmidrule(lr){1-14}
% \multirow{2}{*}{Cyclic}
% & {w/o} & 
% \\
% & {w/} & 
% \\
% \multirow{2}{*}{Mean} 
% & w/o & 
% 81.61 & 79.46 & 76.74 &
% 81.86 & 79.74 &&
% 88.36 & 85.45 & 80.47 &
% 89.09 & 88.93 & 89.12
% \\
% % & w/ & x & 72.34 \\
% & w/ & 
% 81.27 & 81.30 & &
% 81.14 & 81.27 & 82.31 &
% 89.74 & 89.82 & 89.80 &
% 89.46 & 89.49 & 89.39
% \\
\cmidrule(lr){1-17}
& \ding{55} & 
\bf 89.07 & 89.07 & 89.07 &&
91.07 & 91.07 & 91.07 &&
89.80 & 89.80 & 89.80 &&
87.64 & 87.64 & 87.64
\\
\rowcolor{LightCyan} \cellcolor{White}
\multirow{-2}{*}{GPS~\cite{lyzhov2020greedy}} 
& \ding{51} & 
89.01 & \bf 89.22 & \bf 89.19 &&
\bf 91.17 & \bf 91.28 & \bf 91.28 &&
\bf 89.90 & \bf 90.06 & \bf 90.11 &&
\bf 87.66 & \bf 87.71 & \bf 87.75
\\
\cmidrule(lr){1-17}
& \ding{55} & 
88.62 & 87.40 & 83.28 &&
90.78 & 89.09 & 86.29 &&
89.30 & 87.49 & 84.35 &&
\bf 87.77 & 86.63 & 83.75
\\
\rowcolor{LightCyan} \cellcolor{White}
\multirow{-2}{*}{ClassTTA~\cite{shanmugam2021better}} 
& \ding{51} & 
\bf 89.07 & \bf 89.22 & \bf 89.15 &&
\bf 91.19 & \bf 91.28 & \bf 91.36 &&
\bf 89.77 & \bf 89.89 & \bf 90.00 &&
87.53 & \bf 87.66 & \bf 87.53
\\
\cmidrule(lr){1-17}
& \ding{55} & 
88.88 & 88.97 & 87.40 &&
\bf 91.20 & 90.80 & 89.41 &&
\bf 90.05 & 89.93 & 88.19 &&
\bf 87.74 & 87.10 & 86.91
\\
\rowcolor{LightCyan} \cellcolor{White}
\multirow{-2}{*}{AugTTA~\cite{shanmugam2021better}} 
& \ding{51} & 
\bf 89.10 & \bf 89.23 & \bf 89.15 &&
91.13 & \bf 91.28 & \bf 91.28 &&
89.92 & \bf 90.10 & \bf 90.11 &&
87.61 & \bf 87.75 & \bf 87.79
\\
\specialrule{.15em}{.05em}{.05em}
\end{tabular}
}

\end{table*}

%% file: tables/tta-imagenet-expanded-app.tex
\begin{table*}[h!]

\centering
\setlength{\tabcolsep}{4pt}

\captionof{table}{
    \textbf{Comparison to TTA methods on ImageNet with expanded~\cite{shanmugam2021better} TTA policies.} 
    We evaluate on ImageNet under $B_{\texttt{total}} \in \{30, 100, 150\}$ with various model architectures. For each $B_{\texttt{total}}$, we report the top-1 Acc. of baseline TTA methods with and without our learned approach. Whichever is the better one is bolded. The results of our learned procedure are highlighted. 
}
\label{table:tta-imagenet-expanded-app}
\resizebox{\linewidth}{!}{%}
\begin{tabular}{ c c ccc c  ccc c ccc c ccc } 
\specialrule{.15em}{.05em}{.05em}

\multirow{2}{*}{TTA} & 
\multirow{2}{*}{Ours} & 
\multicolumn{3}{c}{ResNext50} && 
\multicolumn{3}{c}{ShuffleNetV2} && 
\multicolumn{3}{c}{Swin} && 
\multicolumn{3}{c}{SwinV2} \\
& & 
$30$ & $100$ & $150$ && 
$30$ & $100$ & $150$ && 
$30$ & $100$ & $150$ && 
30 & 100 & 150 \\
\cmidrule(lr){1-17}
& \ding{55} & 
78.01 & 78.01 & 78.01 &&
70.17 & 70.17 & 70.17 &&
81.35 & 81.35 & 81.35 &&
81.31 & 81.31 & 81.31
\\
\rowcolor{LightCyan} \cellcolor{White}
\multirow{-2}{*}{GPS~\cite{lyzhov2020greedy}} 
& \ding{51} & 
\bf 78.18 & \bf 78.32 & \bf 78.30 &&
\bf 70.44 & \bf 70.35 & \bf 70.14 &&
\bf 81.42 & \bf 81.41 & \bf 81.42 &&
\bf 81.35 & \bf 81.44 & \bf 81.46

\\
\cmidrule(lr){1-17}
& \ding{55} & 
77.03 & 76.70 & 75.45 &&
69.01 & 69.02 & 68.57 &&
80.63 & 80.65 & 80.80 &&
80.56 & 80.54 & 80.57
\\
\rowcolor{LightCyan} \cellcolor{White}
\multirow{-2}{*}{ClassTTA~\cite{shanmugam2021better}} 
& \ding{51} & 
\bf 78.10 & \bf 78.15 & \bf 78.20 &&
\bf 70.36 & \bf 70.15 & \bf 70.07 &&
\bf 81.52 & \bf 81.50 & \bf 81.54 &&
\bf 81.39 & \bf 81.39 & \bf 81.38
\\

\cmidrule(lr){1-17}
& \ding{55} & 
78.09 & 78.13 & 78.08 &&
70.37 & \bf 70.49 & \bf 70.40 &&
81.42 & 81.50 & 81.38 &&
81.28 & 81.41 & 81.25
\\
\rowcolor{LightCyan} \cellcolor{White}
\multirow{-2}{*}{AugTTA~\cite{shanmugam2021better}} 
& \ding{51} & 
\bf 78.14 & \bf 78.23 & \bf 78.28 &&
\bf 70.51 & 70.35 & 70.25 &&
\bf 81.49 & \bf 81.51 & \bf 81.52 &&
\bf 81.40 & \bf 81.42 & \bf 81.42
\\
\specialrule{.15em}{.05em}{.05em}
\end{tabular}
}
\end{table*}

%% file: tables/computation_comparison.tex
\begin{table*}[h]
\small
\centering
\captionof{table}{\textbf{Comparison of computation budget of ResNet18 on ImageNet.} We report the comparison of top-1 Acc. , MACs (G), and the latency (ms/img) between baseline TTA methods w/ and w/o our learned approach.
}
\label{table:computation_comparison}
\setlength{\tabcolsep}{5pt}
\begin{tabular}{ c@{\hskip 5pt}c@{\hskip 8pt}c c r@{\hskip 5pt}r r@{\hskip 5pt}r} 
\specialrule{.15em}{.05em}{.05em}
$B_{\tt total}$  & {TTA} & {Ours} & 
Acc $\uparrow$ & \multicolumn{2}{c}{MACs $\downarrow$} & \multicolumn{2}{c}{Latency $\downarrow$ }
\\
\cmidrule{1-8}
1 & \ding{55} & \ding{55} & 69.76 & 1.8 & (1.0$\times$) & 1.6 & (1.0$\times$)
\\
\cmidrule{1-8}
\multirow{6}{*}{30} & & \ding{55} & 
70.51 & \bf 5.5 & \bf (3.0$\times$) & \bf 4.2 & \bf (2.7$\times$)
\\
\rowcolor{LightCyan} \cellcolor{White} & \cellcolor{White} 
\multirow{-2}{*}{GPS} & \ding{51} & 
\bf 70.74 & 20.8 & (11.4$\times$) & 8.9 & (5.7$\times$)
\\
\cmidrule{2-8}
& & \ding{55} & 
69.09 & \bf 54.7 & \bf (30.0$\times$) & 43.3 & (27.9$\times$)
\\
\rowcolor{LightCyan} \cellcolor{White} & \cellcolor{White} 
\multirow{-2}{*}{ClassTTA} & \ding{51} & 
\bf 70.37 & 69.3 & (38.0$\times$) & \bf 25.2 & \bf (16.2$\times$)
\\
\cmidrule{2-8}
& & \ding{55} & 
70.55 & \bf 54.7 & \bf (30.0$\times$) & 54.0 & (34.8$\times$)
\\
\rowcolor{LightCyan} \cellcolor{White} & \cellcolor{White} 
\multirow{-2}{*}{AugTTA} & \ding{51} & 
\bf 70.75 & 69.3 & (38.0$\times$) & \bf 31.7 & \bf (20.4$\times$)
\\
\cmidrule{1-8}
\multirow{6}{*}{100} &  & \ding{55} &
70.51 & \bf 5.5 & \bf (3.0$\times$) & \bf 4.0 & \bf (2.6$\times$)
\\
\rowcolor{LightCyan} \cellcolor{White} & \cellcolor{White}
\multirow{-2}{*}{GPS} & \ding{51} &
\bf 70.74 & 50.3 & (27.6$\times$) & 24.3 & (15.6$\times$)
\\
\cmidrule{2-8}
& & \ding{55} & 
68.23 & 182.4 & (100.0$\times$) & 155.8 & (100.3$\times$)
\\
\rowcolor{LightCyan} \cellcolor{White} & \cellcolor{White} 
\multirow{-2}{*}{ClassTTA} & \ding{51} &
\bf 70.36 & \bf 167.7 & \bf (91.9$\times$) & \bf 118.8 & \bf (76.5$\times$)
\\
\cmidrule{2-8}
& & \ding{55} & 
70.66 & 182.4 & (100.0$\times$) & 141.4 & (91.0$\times$)
\\
\rowcolor{LightCyan} \cellcolor{White} & \cellcolor{White} 
\multirow{-2}{*}{AugTTA} & \ding{51} &
\bf 70.79 & \bf 167.7 & \bf (91.9$\times$) & \bf 86.2 & \bf (55.5$\times$)
\\
\cmidrule{1-8}
\multirow{6}{*}{150} &  & \ding{55} &
70.51 & \bf 5.5 & \bf (3.0$\times$) & \bf 3.9 & \bf (2.5$\times$)
\\
\rowcolor{LightCyan} \cellcolor{White} & \cellcolor{White}
\multirow{-2}{*}{GPS} & \ding{51} &
\bf 70.69 & 76.1 & (41.7$\times$) & 39.7 & (25.6$\times$)
\\
\cmidrule{2-8}
& & \ding{55} & 
66.40 & 273.6 & (150.0$\times$) & 236.4 & (152.1$\times$)
\\
\rowcolor{LightCyan} \cellcolor{White} & \cellcolor{White} 
\multirow{-2}{*}{ClassTTA} & \ding{51} &
\bf 70.37 & \bf 253.6 & \bf (139.0$\times$) & \bf 159.1 & \bf (102.4$\times$)
\\
\cmidrule{2-8}
& & \ding{55} & 
70.28 & 273.6 & (150.0$\times$) & 244.0 & (157.0$\times$)
\\
\rowcolor{LightCyan} \cellcolor{White} & \cellcolor{White} 
\multirow{-2}{*}{AugTTA} & \ding{51} &
\bf 70.74 & \bf 253.6 & \bf (139.0$\times$) & \bf 131.6 & \bf (84.7$\times$)
\\
\specialrule{.15em}{.05em}{.05em}
\end{tabular}

\end{table*}

%% file: tables/computation_comparison-mnv2.tex
\begin{table*}[h]
\small
\centering
\captionof{table}{\textbf{Comparison of computation budget of MobileNetV2 on ImageNet.} We report the comparison of top-1 Acc. , MACs (G), and the latency (ms/img) between baseline TTA methods w/ and w/o our learned approach.
}
\label{table:computation_comparison-mnv2}
\setlength{\tabcolsep}{5pt}
\begin{tabular}{ c@{\hskip 5pt}c@{\hskip 8pt}c c r@{\hskip 5pt}r r@{\hskip 5pt}r} 
\specialrule{.15em}{.05em}{.05em}
$B_{\tt total}$  & {TTA} & {Ours} & 
Acc $\uparrow$ & \multicolumn{2}{c}{MACs $\downarrow$} & \multicolumn{2}{c}{Latency $\downarrow$ }
\\
\cmidrule{1-8}
1 & \ding{55} & \ding{55} & 
71.88 & 0.3 & (1.0$\times$) & 2.4 & (1.0$\times$)
\\
\cmidrule{1-8}
\multirow{6}{*}{30} & & \ding{55} & 
72.24 & \bf 1.0 & \bf (3.0$\times$) & \bf 7.6 & \bf (3.1$\times$)
\\
\rowcolor{LightCyan} \cellcolor{White} & \cellcolor{White} 
\multirow{-2}{*}{GPS} & \ding{51} & 
\bf 72.37 & 3.5 & (10.7$\times$) & 11.8 & (4.9$\times$)
\\
\cmidrule{2-8}
& & \ding{55} & 
70.58 & \bf 9.8 & \bf (30.0$\times$) & 87.0 & (36.0$\times$)
\\
\rowcolor{LightCyan} \cellcolor{White} & \cellcolor{White} 
\multirow{-2}{*}{ClassTTA} & \ding{51} & 
\bf 71.44 & 11.6 & (35.6$\times$) & \bf 40.9 & \bf (16.9$\times$)
\\
\cmidrule{2-8}
& & \ding{55} & 
72.33 & \bf 9.8 & \bf (30.0$\times$) & 86.2 & (35.7$\times$)
\\
\rowcolor{LightCyan} \cellcolor{White} & \cellcolor{White} 
\multirow{-2}{*}{AugTTA} & \ding{51} & 
\bf 72.41 & 11.7 & (35.6$\times$) & \bf 46.0 & \bf (19.1$\times$)
\\
\cmidrule{1-8}
\multirow{6}{*}{100} &  & \ding{55} &
72.24 & \bf 1.0 & \bf (3.0$\times$) & \bf 7.4 & \bf (3.1$\times$)
\\
\rowcolor{LightCyan} \cellcolor{White} & \cellcolor{White}
\multirow{-2}{*}{GPS} & \ding{51} &
\bf 72.61 & 9.2 & (28.1$\times$) & 38.9 & (16.1$\times$)
\\
\cmidrule{2-8}
& & \ding{55} & 
69.97 & 32.8 & (100.0$\times$) & 260.6 & (108.0$\times$)
\\
\rowcolor{LightCyan} \cellcolor{White} & \cellcolor{White} 
\multirow{-2}{*}{ClassTTA} & \ding{51} &
\bf 71.68 & \bf 30.7 & \bf (93.6$\times$) & \bf 140.4 & \bf (58.2$\times$)
\\
\cmidrule{2-8}
& & \ding{55} & 
72.42 & 32.8 & (100.0$\times$) & 270.9 & (112.3$\times$)
\\
\rowcolor{LightCyan} \cellcolor{White} & \cellcolor{White} 
\multirow{-2}{*}{AugTTA} & \ding{51} &
\bf 72.62 & \bf 30.7 & \bf (93.6$\times$) & \bf 120.9 & \bf (50.1$\times$)
\\
\cmidrule{1-8}
\multirow{6}{*}{150} &  & \ding{55} &
72.24 & \bf 1.0 & \bf (3.0$\times$) & \bf 8.5 & \bf (3.5$\times$)
\\
\rowcolor{LightCyan} \cellcolor{White} & \cellcolor{White}
\multirow{-2}{*}{GPS} & \ding{51} &
\bf 72.58 & 14.6 & (44.5$\times$) & 69.5 & (28.8$\times$)
\\
\cmidrule{2-8}
& & \ding{55} & 
67.81 & 49.1 & (150.0$\times$) & 481.8 & (199.7$\times$)
\\
\rowcolor{LightCyan} \cellcolor{White} & \cellcolor{White} 
\multirow{-2}{*}{ClassTTA} & \ding{51} &
\bf 71.63 & \bf 48.6 & \bf (148.4$\times$) & \bf 235.5 & \bf (97.6$\times$)
\\
\cmidrule{2-8}
& & \ding{55} & 
72.46 & 49.1 & (150.0$\times$) & 435.9 & (180.6$\times$)
\\
\rowcolor{LightCyan} \cellcolor{White} & \cellcolor{White} 
\multirow{-2}{*}{AugTTA} & \ding{51} &
\bf 72.58 & \bf 48.6 & \bf (148.4$\times$) & \bf 246.9 & \bf (102.3$\times$)
\\
\specialrule{.15em}{.05em}{.05em}
\end{tabular}

\end{table*}

%% file: tables/tta-imagenet-expanded-non_learn.tex
\begin{table*}[h!]
% \small
\centering
\setlength{\tabcolsep}{4pt}
\captionof{table}{
    \textbf{Comparison to non-learned TTA methods on ImageNet with expanded~\cite{shanmugam2021better} TTA policies.} 
    We evaluate on ImageNet under $B_{\texttt{total}} \in \{30, 100, 150\}$ with various model architectures. For each $B_{\texttt{total}}$, we report the top-1 Acc. of non-learned TTA methods with and without our non-learned approach. Whichever is the better one is bolded. The results of our non-learned procedure are highlighted. 
}
\label{table:tta-imagenet-expanded-non_learn}
\resizebox{\linewidth}{!}{%}
\begin{tabular}{ c c c ccc c ccc c ccc c ccc } 
\specialrule{.15em}{.05em}{.05em}

& \multirow{2}{*}{Method} & 
\multicolumn{3}{c}{ResNet18} && 
\multicolumn{3}{c}{ResNet50} && 
\multicolumn{3}{c}{MobileNetV2} && 
\multicolumn{3}{c}{InceptionV3}\\
& & 
$30$ & $75$ & $150$ && 
$30$ & $75$ & $150$ && 
$30$ & $75$ & $150$ && 
$30$ & $75$ & $150$ 
\\
\cmidrule(lr){1-17} 
& MeanTTA & 
67.23 & 68.37 & 67.81 &&
73.88 & 74.77 & 74.30 &&
68.72 & 69.98 & 69.44 &&
70.46 & 70.71 & 70.51
\\
& MaxTTA & 
66.31 & 66.07 & 65.60 &&
71.79 & 71.52 & 71.12 &&
68.20 & 67.92 & 67.32 &&
64.95 & 65.21 & 64.98
\\
\rowcolor{LightCyan} \cellcolor{White}
% \multirow{-2}{*}{Mean}
& Ours &
\bf 70.27 & -- & -- &&
\bf 76.54 & -- & -- &&
\bf 72.35 & -- & -- &&
\bf 71.45 & -- & -- 
\\
% \cmidrule(lr){1-17}

% \\
% \rowcolor{LightCyan} \cellcolor{White}
% % \multirow{-2}{*}{Max} 
% & Ours & 
% \bf 70.27 & -- & -- &&
% \bf 76.54 & -- & -- &&
% \bf 72.35 & -- & -- &&
% \bf 71.45 & -- & --
% \\
\specialrule{.15em}{.05em}{.05em}
\end{tabular}
}
\end{table*}

%% file: tables/tta-imagenet-standard-non_learn.tex
\begin{table*}[h!]
% \small
\centering
\setlength{\tabcolsep}{4pt}
\captionof{table}{
    \textbf{Comparison to non-learned TTA methods on ImageNet with standard~\cite{shanmugam2021better} TTA policies.} 
    We evaluate on ImageNet under $B_{\texttt{total}} \in \{30, 100, 150\}$ with various model architectures. For each $B_{\texttt{total}}$, we report the top-1 Acc. of non-learned TTA methods with and without our non-learned approach. Whichever is the better one is bolded. The results of our non-learned procedure are highlighted. 
}
\label{table:tta-imagenet-standard-non_learn}
\resizebox{\linewidth}{!}{%}
\begin{tabular}{ c c c ccc c ccc c ccc c ccc } 
\specialrule{.15em}{.05em}{.05em}

\multirow{2}{*}{TTA} & \multirow{2}{*}{Ours} & 
\multicolumn{3}{c}{ResNet18} && 
\multicolumn{3}{c}{ResNet50} && 
\multicolumn{3}{c}{MobileNetV2} && 
\multicolumn{3}{c}{InceptionV3}\\
& & 
$30$ & $75$ & $150$ && 
$30$ & $75$ & $150$ && 
$30$ & $75$ & $150$ && 
$30$ & $75$ & $150$ 
\\
\cmidrule(lr){1-17} 
& \ding{55} & 
\bf 70.90 & 70.38 & 68.06 &&
76.65 & 75.99 & 73.57 &&
\bf 72.65 & 72.00 & 68.98 &&
\bf 71.99 & \bf 71.61 & 70.44
\\
\rowcolor{LightCyan} \cellcolor{White}
\multirow{-2}{*}{MeanTTA}
& \ding{51} &
70.70 & \bf 70.84 & \bf 70.81 &&
\bf 76.71 & \bf 76.76 & \bf 76.83 &&
72.61 & \bf 72.57 & \bf 72.71 &&
71.44 & 71.58 & \bf 71.98
\\

\cmidrule(lr){1-17}
& \ding{55} & 
70.02 & 69.70 & 68.39 &&
76.24 & 75.62 & 72.86 &&
72.13 & 71.79 & 70.10 &&
\bf 71.42 & 69.93 & 67.27
\\
\rowcolor{LightCyan} \cellcolor{White}
\multirow{-2}{*}{MaxTTA} 
& \ding{51} & 
\bf 70.24 & \bf 70.45 & \bf 70.46 &&
\bf 76.49 & \bf 76.51 & \bf 76.66 &&
\bf 72.33 & \bf 72.43 & \bf 72.54 &&
71.10 & \bf 71.33 & \bf 71.76
\\
\specialrule{.15em}{.05em}{.05em}
\end{tabular}
}
\end{table*}

%% file: tables/antialias-cnn.tex
\begin{table*}[h!]

\centering
\setlength{\tabcolsep}{4pt}
\captionof{table}{
    \textbf{Performance on the pre-trained weights of anti-aliased CNN~\cite{zhang2019making}.}
    We consider various antialiased CNN backbones with and without our procedure under different $B_{\texttt{ours}} \in \{4, 10\}$.
    The \ding{55} indicates the experiments without ours by setting $B_{\texttt{ours}} = 1$.
    We report the top-1 Acc. on ImageNet. The results w/ our non-learned procedure are highlighted.
    Our procedure makes improvements on all antialiased CNN backbones.
}
\label{table:antialias-cnn}
\resizebox{\linewidth}{!}{%}
\begin{tabular}{ ccc c ccc c ccc c ccc c ccc } 
\specialrule{.15em}{.05em}{.05em}
\multicolumn{3}{c}{ResNet18} && 
\multicolumn{3}{c}{ResNet34} && 
\multicolumn{3}{c}{ResNet50} && 
\multicolumn{3}{c}{WideResNet50} && 
\multicolumn{3}{c}{MobileNetV2-050} \\
\ding{55} & 4 & 10 &&
\ding{55} & 4 & 10 &&
\ding{55} & 4 & 10 &&
\ding{55} & 4 & 10 &&
\ding{55} & 4 & 10
\\
\cmidrule(lr){1-19}
71.67 & \ours \underline{71.74} & \ours \bf 71.88 &&
74.60 & \ours \underline{74.75} & \ours \bf 74.90 &&
77.41 & \ours \underline{77.43} & \ours \bf 77.53 &&
78.70 & \ours \underline{78.76} & \ours \bf 78.91 &&
72.72 & \ours \underline{72.98} & \ours \bf 73.14
\\
\specialrule{.15em}{.05em}{.05em}
\end{tabular}
}
\end{table*}

% Model 	Baseline 	W/ antialias-CNN 	W/ antialias-CNN + Ours (B=4) 	W/ antialias-CNN + Ours (B=10)
% ResNet-18 	69.74 	71.67 	71.74 	71.88
% ResNet-34 	73.30 	74.60 	74.75 	74.90
% ResNet-50 	76.16 	77.41 	77.43 	77.53
% WideResNet-50 	78.47 	78.70 	78.76 	78.91
% MobileNetV2-050 	71.88 	72.72 	72.98 	73.14

%% file: tables/tta-standard.tex
\begin{table*}[h!]
\small
\centering
\setlength{\tabcolsep}{4pt}
\captionof{table}{
    \textbf{Details of our modified standard policy.} We modified {\tt standard}~\cite{shanmugam2021better} TTA policy to cover budget up to $B_{\tt tta}=190$.
}
\label{table:tta-standard}
\resizebox{\linewidth}{!}{%}
\begin{tabular}{ l l l l  } 
\specialrule{.15em}{.05em}{.05em}
Augmentation & \# Aug. & Range of $p$ & Description
\\
\cmidrule{1-4}
{\tt FlipLR} & 2 & 
\begin{tabular}[t]{l}
\tt{False}, \tt{True}
\end{tabular}
& 
\begin{tabular}[t]{l}
Horizontally flip the image if $p=$ {\tt True}.
\end{tabular}
\\
\cmidrule{1-4}
{\tt Scale} & 19 & 
\begin{tabular}[t]{l}
1.00, 1.04, 1.10, 0.98, 0.92, 0.86, 0.80, \\ 
0.74, 0.68, 0.62, 0.56, 0.50, 0.44, 0.38, \\ 
0.32, 0.26, 0.20, 0.14, 0.08
\end{tabular}
&
\begin{tabular}[t]{l}
Scale the image by a ratio of $p$.
\end{tabular}
\\
\cmidrule{1-4}
{\tt FiveCrop} & 5 & 
\begin{tabular}[t]{l}
{\tt Center}, {\tt LeftTop}, {\tt LeftBottom}, \\ 
{\tt RightTop}, {\tt RightBottom}
\end{tabular}
& 
\begin{tabular}[t]{l}
Crop from the $p$ (center or \\
one of the corners) of the image.
\end{tabular}
\\
\specialrule{.15em}{.05em}{.05em}
\end{tabular}
}
\end{table*}

%% file: tables/tta-expanded.tex
\begin{table*}[h!]
\small
\centering
\setlength{\tabcolsep}{4pt}
\captionof{table}{
    \textbf{Details of the our modified expanded policy.} We modified {\tt expanded}~\cite{shanmugam2021better} TTA policy to cover budget up to $B_{\tt tta}=8778$.
}
\label{table:tta-expanded}
\resizebox{\linewidth}{!}{%}
\begin{tabular}{ l l l l  } 
\specialrule{.15em}{.05em}{.05em}
Augmentation & \# Aug. & Range of $p$ & Description
\\
\cmidrule{1-4}
{\tt Identity} & 1 & 
\begin{tabular}[t]{l}
--
\end{tabular}
& 
\begin{tabular}[t]{l}
Return the original image.
\end{tabular}
\\
\cmidrule{1-4}
{\tt FlipLR} & 1 & 
\begin{tabular}[t]{l}
--
\end{tabular}
& 
\begin{tabular}[t]{l}
Horizontally flip the image.
\end{tabular}
\\

\cmidrule{1-4}
{\tt FlipUD} & 1 & 
\begin{tabular}[t]{l}
--
\end{tabular}
& 
\begin{tabular}[t]{l}
Vertically flip the image.
\end{tabular}
\\

\cmidrule{1-4}
{\tt Invert} & 1 & 
\begin{tabular}[t]{l}
--
\end{tabular}
& 
\begin{tabular}[t]{l}
Invert the colors of the image.
\end{tabular}
\\

\cmidrule{1-4}
{\tt PIL\_Blur} & 1 & 
\begin{tabular}[t]{l}
--
\end{tabular}
& 
\begin{tabular}[t]{l}
Blur the image using the \\
{\tt PIL.ImageFilter.BLUR} kernel.
\end{tabular}
\\

\cmidrule{1-4}
{\tt PIL\_Smooth} & 1 & 
\begin{tabular}[t]{l}
--
\end{tabular}
& 
\begin{tabular}[t]{l}
Smooth the image using the \\
{\tt PIL.ImageFilter.Smooth} kernel.
\end{tabular}
\\

\cmidrule{1-4}
{\tt AutoContrast} & 1 & 
\begin{tabular}[t]{l}
--
\end{tabular}
& 
\begin{tabular}[t]{l}
Maximize the contrast of the image.
\end{tabular}
\\

\cmidrule{1-4}
{\tt Equalize} & 1 & 
\begin{tabular}[t]{l}
--
\end{tabular}
& 
\begin{tabular}[t]{l}
Equalize the histogram of the image.
\end{tabular}
\\

\cmidrule{1-4}
{\tt Posterize} & 4 & 
\begin{tabular}[t]{l}
1, 2, 3, 4
\end{tabular}
& 
\begin{tabular}[t]{l}
Posterize the image by reducing $p$ RGB bits.
\end{tabular}
\\

\cmidrule{1-4}
{\tt Rotate} & 10 & 
\begin{tabular}[t]{l}
{\tt linspace}(-30, 30, 10)
\end{tabular}
& 
\begin{tabular}[t]{l}
Rotate the image by $p$ degree.
\end{tabular}
\\

\cmidrule{1-4}
{\tt CropBilinear} & 10 & 
\begin{tabular}[t]{l}
\{1, 2, $\cdots$, 10\}
\end{tabular}
& 
\begin{tabular}[t]{l}
Crop by (224-$p$)px and then resize the image.
\end{tabular}
\\

\cmidrule{1-4}
{\tt Solarize} & 10 & 
\begin{tabular}[t]{l}
{\tt linspace}(0, 1, 10)
\end{tabular}
& 
\begin{tabular}[t]{l}
Solarize the image by a threshold of $p$.
\end{tabular}
\\

\cmidrule{1-4}
{\tt Contrast} & 10 & 
\begin{tabular}[t]{l}
{\tt linspace}(0.1, 1.9, 10)
\end{tabular}
& 
\begin{tabular}[t]{l}
Adjust the contrast of the image \\ by a contrast factor of $p$.
\end{tabular}
\\

\cmidrule{1-4}
{\tt Saturation} & 10 & 
\begin{tabular}[t]{l}
{\tt linspace}(0.1, 1.9, 10)
\end{tabular}
& 
\begin{tabular}[t]{l}
Adjust the saturation of the image \\ by a saturation factor of $p$.
\end{tabular}
\\

\cmidrule{1-4}
{\tt Brightness} & 10 & 
\begin{tabular}[t]{l}
{\tt linspace}(0.1, 1.9, 10)
\end{tabular}
& 
\begin{tabular}[t]{l}
Adjust the brightness of the image \\ by a brightness factor of $p$.
\end{tabular}
\\

\cmidrule{1-4}
{\tt Sharpness} & 10 & 
\begin{tabular}[t]{l}
{\tt linspace}(0.1, 1.9, 10)
\end{tabular}
& 
\begin{tabular}[t]{l}
Adjust the sharpness of the image \\ by a sharpness factor of $p$.
\end{tabular}
\\

\cmidrule{1-4}
{\tt ShearX} & 10 & 
\begin{tabular}[t]{l}
{\tt linspace}(-0.3, 0.3, 10)
\end{tabular}
& 
\begin{tabular}[t]{l}
Shear the image along the $x$-axis \\ by $\tan^{-1}(p)$ radians.
\end{tabular}
\\

\cmidrule{1-4}
{\tt ShearY} & 10 & 
\begin{tabular}[t]{l}
{\tt linspace}(-0.3, 0.3, 10)
\end{tabular}
& 
\begin{tabular}[t]{l}
Shear the image along the $y$-axis \\ by $\tan^{-1}(p)$ radians.
\end{tabular}
\\

\cmidrule{1-4}
{\tt TranslateX} & 10 & 
\begin{tabular}[t]{l}
{\tt linspace}(-9, 9, 10)
\end{tabular}
& 
\begin{tabular}[t]{l}
Translate the image along the $x$-axis by $p$ px.
\end{tabular}
\\

\cmidrule{1-4}
{\tt TranslateY} & 10 & 
\begin{tabular}[t]{l}
{\tt linspace}(-9, 9, 10)
\end{tabular}
& 
\begin{tabular}[t]{l}
Translate the image along the $y$-axis by $p$ px.
\end{tabular}
\\

\cmidrule{1-4}
{\tt Cutout} & 10 & 
\begin{tabular}[t]{l}
{\tt linspace}(2, 20, 10)
\end{tabular}
& 
\begin{tabular}[t]{l}
Randomly mask out $p$ px by $p$ px from the image.
\end{tabular}\\
\specialrule{.15em}{.05em}{.05em}
\end{tabular}
}

\end{table*}